%% file: main.tex
\documentclass[10pt,twocolumn,letterpaper]{article}

\usepackage[pagenumbers]{iccv} %

\input{preamble}

\definecolor{iccvblue}{rgb}{0.21,0.49,0.74}
\usepackage[pagebackref,breaklinks,colorlinks,allcolors=iccvblue]{hyperref}

\include{command_notation}
\usepackage{graphicx}
\usepackage{tabularx}
\usepackage{amsmath}

\def\paperID{1699} %
\def\confName{ICCV}
\def\confYear{2025}

\title{TwoSquared: 4D Generation from 2D Image Pairs}

\author{
Lu Sang$^{*1,2}$, Zehranaz Canfes$^{*1}$, Dongliang Cao$^{3}$\\
Riccardo Marin$^{1,2}$, Florian Bernard$^{3}$, Daniel Cremers$^{1,2}$ \\
$^{1}$Technical University of Munich, $^{2}$Munich Center of Machine Learning\\
$^{3}$University of Bonn \\
}

\begin{document}

\twocolumn[{%
\renewcommand\twocolumn[1][]{#1}%
\maketitle

\begin{center}
    \centering
    \begin{overpic}[trim=0cm 0.1cm 0cm 0.3cm,clip, width=\linewidth]{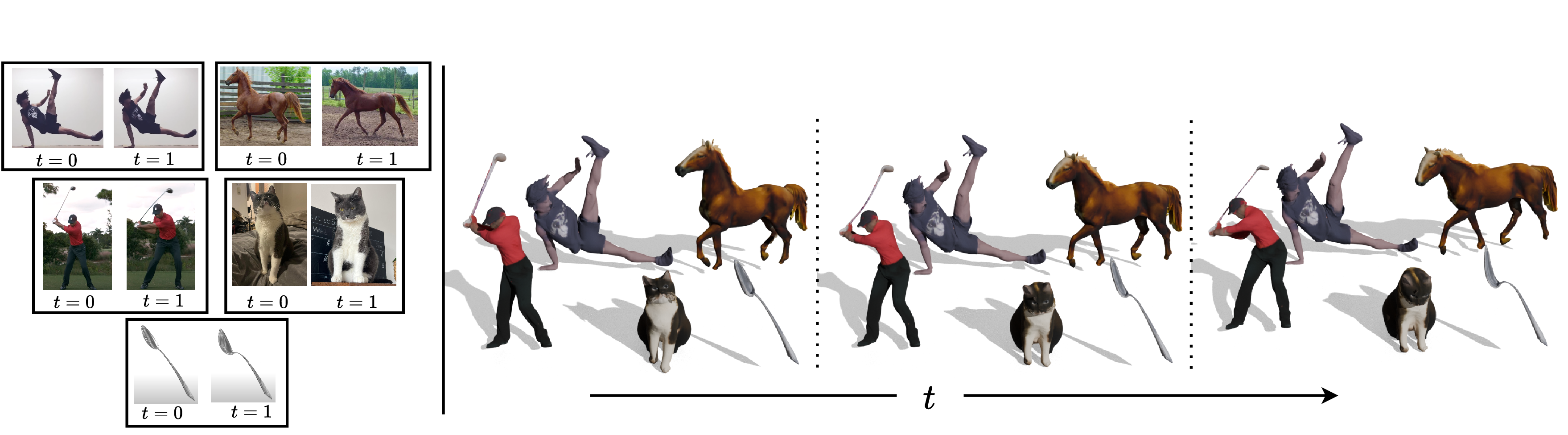}
			\put(5, 25){Input: Image Pairs}
            \put(53, 25){Output: Generated 4D Sequences}
            \end{overpic}
    \captionof{figure}{\methodname takes a pair of 2D images representing the initial and final states of an object as input and generates texture-consistent, geometry-consistent 4D continuous sequences. It is designed to be robust to varying input quality, operating without the need for predefined templates or object-class priors. This adaptability enables greater flexibility in processing diverse images while maintaining structural integrity and visual coherence throughout the generated sequences. As demonstrated, our approach effectively handles humans, animals, and inanimate objects.}\label{fig:teaser}
\end{center}%

}]

\let\thefootnote\relax\footnote{$^{*}$ These authors contributed equally.}

\begin{abstract}

\noindent Despite the astonishing progress in generative AI, 4D dynamic object generation remains an open challenge. With limited high-quality training data and heavy computing requirements, the combination of hallucinating unseen geometry together with unseen movement poses great challenges to generative models. In this work, we propose \methodname as a method to obtain a 4D physically plausible sequence starting from only two 2D RGB images corresponding to the beginning and end of the action. Instead of directly solving the 4D generation problem, \methodname decomposes the problem into two steps: 1) an image-to-3D module generation based on the existing generative model trained on high-quality 3D assets, and 2) a physically inspired deformation module to predict intermediate movements. To this end, our method does not require templates or object-class-specific prior knowledge and can take in-the-wild images as input. In our experiments, we demonstrate that \methodname is capable of producing texture-consistent and geometry-consistent 4D sequences only given 2D images.
\end{abstract}

\input{sec/1_intro}

\input{sec/2_related_work}

\input{sec/3_method}

\input{sec/4_experiment}
\input{sec/5_conclusion}

{
    \small
    \bibliographystyle{ieeenat_fullname}
    \bibliography{main}
}

\input{sec/X_suppl}

\end{document}

%% file: preamble.tex
\newcommand{\methodname}{TwoSquared\xspace}

\usepackage{pifont}
\usepackage{amssymb}
\usepackage{amsthm}
\usepackage{mathtools}%

\usepackage{makecell}

\usepackage{booktabs} %

\newcommand{\rom}[1]{\uppercase\expandafter{\romannumeral #1\relax}}

\usepackage{calc}

\usepackage{colortbl}
\usepackage{tabulary}
\usepackage{tabularx}
\usepackage{etoolbox}
\usepackage{multirow}
\usepackage{cuted}
\usepackage[percentage]{overpic} 
\usepackage{pgfplots}
\pgfplotsset{compat=newest}%

\usepackage{svg}

\newcommand{\vet}[1]{\mathbf{#1}}

\newcommand{\V}{\mathcal{V}}

\newcommand{\F}{\mathcal{F}}

\newcommand{\dd}{\mathrm{d}}
\newcommand{\norm}[1]{\left\lVert#1\right\rVert}

\newcommand{\dom}{\Omega}
\newcommand \eq[1]{\begin{equation}\begin{aligned}#1\end{aligned}\end{equation}}
\DeclareMathOperator{\tr}{Tr}
\newcommand{\SAstd}{$\text{SA}\sigma$}
\DeclareMathOperator{\dist}{dist}

\newcommand*{\inparagraph}[1]{\noindent\textbf{#1}\hspace{0.5em}}

%% file: sec/1_intro.tex
\section{Introduction}
\label{sec:intro}
4D generation,
which extends traditional 3D modeling by incorporating temporal dynamics, plays a crucial role in numerous applications across computer vision, such as character animation~\cite{hu2023animate}, virtual reality (VR)~\cite{virtual2019reality, schmalstieg2016augmented}, robotics, and autonomous systems~\cite{wimbauer2023behind}.
Unlike static 3D models, 4D generation captures motion and deformation over time, enabling a more comprehensive representation of dynamic objects. However, most existing 4D generation methods rely on highly constrained input data, such as synchronized multi-view video recordings~\cite{oswald2014generalized,pumarola2021d, yan2023nerf}.\\
Our goal is to recover the 4D dynamics of an object from the simplest possible input: an initial and a final 2D RGB image. This approach offers greater control over deformation, as the transformation between the source and target states is explicitly defined by the two images. We believe this setup is more practical in real-world applications, where motion often needs to be directed toward a specific target pose or gesture. 
However, this poses significant challenges, including:
\begin{enumerate*}[label=(\roman*)]
    \item Ensuring texture and geometric consistency throughout the generated 4D sequence.
    \item Producing realistic deformations that match the desired target pose while preserving structural integrity.
\end{enumerate*}
 Moreover, we aim to solve this without relying on any category-specific template~\cite{varen2024, li2024dessie, zhu2025ar4d} or object-class-specific prior knowledge~\cite{curreli2025nonisotropic, Zhang_2024_CVPR}, making the method generalizable across different object types.\\
In recent years, deep learning has made great progress on generative tasks, similar in 3D/4D generation tasks~\cite{xue2024human3diffusion, Zhang_2024_CVPR, zhang2024clay}. The standard approach considers the input image as a condition of the generation process and hallucinates four or more views of the same object~\cite{sun2024dimensionx, li2025dreammesh4d}. Since these generated images are often inconsistent with multiple views, recent methods adopt different techniques to incorporate 3D priors, which generally rely on an available template for the class\cite{zhang2024sifu}, or by combining the views into a common 3D representation \cite{xue2024human3diffusion, xu2024grm}.\\
Despite these advancements, synthesizing temporally coherent moving objects remains an open challenge. While direct video synthesis has produced visually impressive results, closer inspection reveals fundamental physical inaccuracies~\cite{bansal2024videophy,chen2025physical}.\\
In this work, we leverage generative AI just for the 3D generation of the initial and final frame and use a physically-inspired optimization to interpolate between the generated 3D objects. Our method, \methodname, as shown in~\cref{fig:teaser}, takes a pair of images as input and aims to generate a 4D sequence that is texture and geometry-consistent. Moreover, our method can generate physically plausible 4D continuous sequences. In contrast, state-of-the-art 4D generation methods typically rely on temporal input data (e.g., full video sequences) and employ cross-attention mechanisms across temporal, spatial and multi-view dimensions~\cite{zhang20254diffusion}, which is computationally expensive and lacks explicit physical control.\\
To address these limitations, we propose \methodname, a novel approach that combines 3D generation with physically inspired optimization to obtain a physically realistic 4D sequence given a pair of input images. To this end, our method is capable of generating 4D sequences that maintain texture and geometric consistency while ensuring physically plausible motion dynamics. \\
To summarize our contribution:
\begin{itemize}
    \item We propose \methodname, a novel \emph{4D generation method that requires only two 2D frames as input}. Our approach is efficient, template-free, and imposes no restrictions on the types of objects it can handle;
    \item \methodname generates physically plausible intermediate shapes while maintaining texture consistency throughout the 4D sequence;
    \item Our method is efficient and lightweight, does not depend on high-quality multi-view video data, and allows for an arbitrary frame-rate generation without retraining;
    \item We demonstrate superior performance over state-of-the-art methods, open new possibilities for applications and commit to releasing our code to facilitate future research.
\end{itemize}
Our code is available on \url{https://sangluisme.github.io/TwoSquared/}.

%% file: sec/2_related_work.tex
\section{Related Works}
\label{sec:related_works}

\subsection{Single Image to 3D Generation}
Generating high quality 3D representations (e.g.\ meshes~\cite{liu2023one2345,wu2024unique3d,xu2024instantmesh,long2024wonder3d}, NeRF~\cite{mildenhall2021nerf,gu2023nerfdiff,melas2023realfusion}, 3DGS~\cite{kerbl20233d,szymanowicz2024splatter,zou2024triplane}) from a single image is an emerging area.
Earlier works~\cite{shen2023anything,tang2023make,xu2023neurallift,qian2023magic123} attempted to distill prior knowledge from 2D
image diffusion models~\cite{rombach2022high,podell2023sdxl} to create 3D models from text or images via Score Distillation Sampling~\cite{poole2022dreamfusion}. Despite their compelling results, these methods suffer from two main limitations, efficiency and consistency, due to per-instance optimization and single-view ambiguity~\cite{long2024wonder3d}. To improve the efficiency, later works~\cite{long2024wonder3d,liu2023zero,shi2023zero123++,liu2023syncdreamer} seperated the generation process into multi-view generation and 3D reconstruction. To generate consistent multi-view images, pretrained 2D image diffusion models are finetuned using large 3D object datasets~\cite{deitke2023objaverse,deitke2024objaverse}. Despite the visual appealing results, the reconstructed meshes often fail to meet the requirements for downstream tasks (e.g.\ shape animation). To guarantee high-quality mesh generation, most recent methods~\cite{zhang20233dshape2vecset,zhang2024clay,zhao2024michelangelo,hunyuan3d2025} discard the use of pre-trained 2D image diffusion models but train a 3D shape generation model from scratch, resulting in much more detailed geometry generation. Therefore, we utilize the most recent single image to mesh generation method~\cite{hunyuan3d2025} in our paper.   

\begin{figure*}[t]
    \centering
    \includegraphics[width=.9\linewidth]{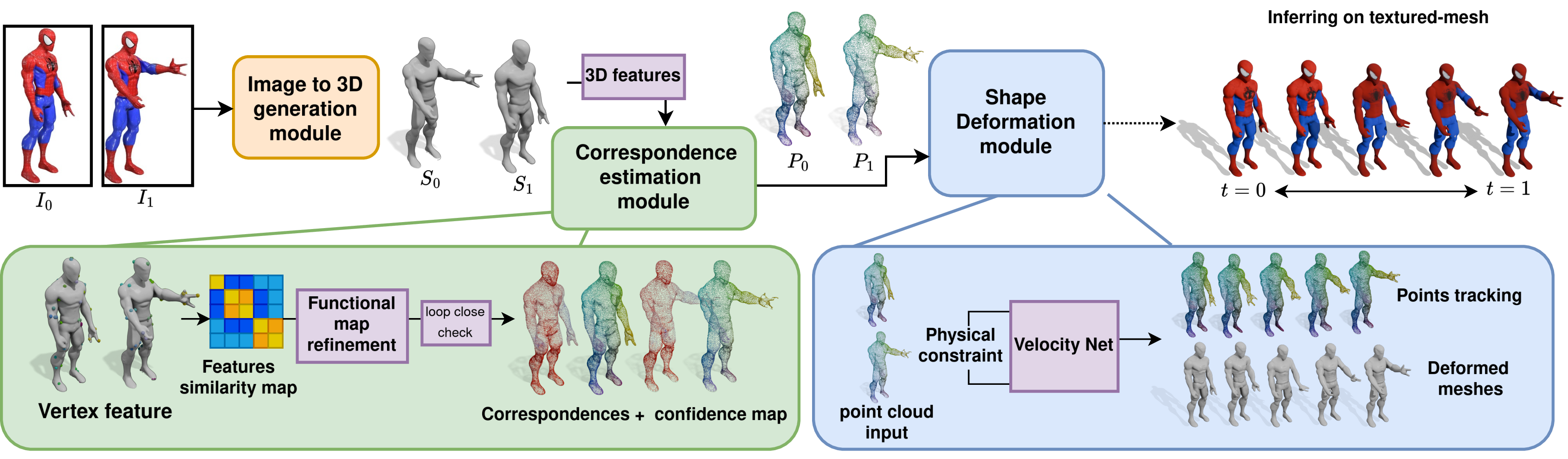}
    \caption{\textbf{Pipeline of TwoSquared}: TwoSquared processes two input images through an image-to-3D generation block, producing two 3D meshes. We then extract per-vertex features and compute a cosine similarity map, which is refined using a functional map module and a close loop check module to obtain point-to-point correspondences and a confidence map. These registered points are then fed into our shape deformation module, where we model their trajectory of the deformed point cloud. During the inference time, we can directly infer the generated textured mesh from $I_0$ to obtain the 4D sequence.
}\label{fig:pipeline}
\vspace*{-2mm}
\end{figure*}

\subsection{4D Generation}
4D generation methods aim to synthesize dynamic, time-evolving 3D representations of objects or scenes by capturing both spatial and temporal information. They can be broadly categorized by their technique –template-free or template-based– or by their input types, which can be videos, 3D data, or 2D images.

\inparagraph{Template-based generation.} Some methods rely on predefined templates or parametric models such as human bodies~\cite{curreli2025nonisotropic}, face, or animals~\cite{li2024dessie, varen2024}. They either directly deform the template using control points, or only create synthetic 4D sequences. However, these methods are limited by the templates they use and are not adaptable to other categories of objects.

\inparagraph{Video-to-4D generation.} One can generate 4D sequences using videos. Methods such as ~\cite{zhu2025ar4d, li2025dreammesh4d, zhang20244diffusion, liang2024diffusion4d} take a video as input, then treat each frame individually to generate the 3D shapes, which are then combined into 4D sequences. These methods often heavily rely on the amount of the training data and the input requirement is high. 

\inparagraph{3D-to-4D generation.} Next, some methods rely directly on 3D input in form of point clouds~\cite{cao2024motion2vecsets, vu2024rfnet}. While they can successfully reconstruct dynamic 4D shapes, they require available point clouds or meshes of the same object. Additionally, \cite{cao2024motion2vecsets} is limited to interpolating only within the set of trained objects, significantly restricting its applicability in more diverse or real-world scenarios. To overcome the limitation of requiring full 3D sequences as inputs, some methods take 3D keyframes as input, such as~\cite{sang20254deform}. They attempt to handle noisy correspondences between keyframes by optimizing the deformation fields across them. These approaches improve robustness in dynamic scenes, where correspondences may be ambiguous or unreliable.

\inparagraph{2D-to-4D generation.} A more challenging task of generating 4D sequences from 2D inputs is addressed by some methods such as ~\cite{sun2024dimensionx, lin2024phys4dgen}. These methods either leverage diffusion models or text prompts to generate a sequence of images and then generate 4D sequences from these images. Consequently, their results is limited by the backbone diffusion or language models.  Moreover, these methods are highly uncontrollable and mostly work on synthetic images.

%% file: sec/3_method.tex
\section{Method}
\label{sec:method}

\inparagraph{Motivation and overview.} A 4D mesh sequence that captures the motion of an object must preserve texture consistency, geometric coherence, and physical plausibility. To ensure both ease of use and practical applicability in real-world scenarios, we design our method with these principles in mind. As illustrated in~\cref{fig:pipeline}, our approach takes two object-centric images as input and generates a temporally consistent 4D deformation between them. The method consists of three components: 
\begin{enumerate*}[label=(\roman*)]
    \item 3D Generation Block – Generates the 3D shapes from the input images.
    \item Vertex Registration Block – Finds correspondences between generated shapes to offer guidance to the follow-up deformation.
    \item Shape Deformation Block – Ensures smooth and realistic transitions across the 4D sequence.
\end{enumerate*}
Our method takes a simple image pair as input and generates a 4D sequence. Instead of producing discrete keyframe meshes, it constructs a continuous deformation field, allowing for the generation of 4D sequences at arbitrary frame rates.

\subsection{Image-to-3D Generation Block}
Our pipeline is flexible and can be integrated with any 3D generation method. In this paper, we adopt the latest Hunyuan3D~\cite{hunyuan3d2025} as our image-to-3D generation block. The reason is that it can generate 3D shapes from various categories, including humans, animals, and sketches, with satisfying geometry details and high-resolution texture maps. Their method can effectively generate meshes with fine-grained geometric details by training a large latent diffusion model on large high-quality 3D assets. Alongside the detailed geometry, the method also generates high-resolution texture maps using their texture-painting module. In our pipeline, by feeding the network with a pair of keyframe images $\mathcal{I} = \{I_i\}, \ i\in\{0, 1\}$, we obtain 3D meshes $\mathcal{S} = \{S_i\}, \  i\in\{0,1\}$ of keyframes. These keyframe meshes are then fed into our vertex-registration module to obtain sparse correspondences. 

\subsection{Vertex Registration Block}
After obtaining the 3D meshes from the image-to-3D module, we render $N=100$ depth and normal images and follow the pipeline of Diff3F~\cite{dutt2024diff} to obtain the diffusion features of the meshes. 
Then, these per-pixel features are unprojected on the 3D meshes. For each mesh $i$ with $v_i$ vertices, we obtain a feature matrix with dimension $\F_i \in\mathbb{R}^{v_i\times f}$, where $f=2048$ is the feature dimension. After that, for mesh $S_i$ and $S_j$, we compute the cosine similarity matrix $M_{ij}\in\mathbb{R}^{v_i \times v_j}$ from their feature matrices $\F_i, \F_j$.
Thus, we can obtain point-to-point correspondences through $M_{ij}$. However, the correspondences obtained directly from $M_{ij}$ are noisy and inaccurate, since it only considers the information of each individual point, leading to errors such as multiple points in $S_i$ being assigned the same point in $S_j$. To solve this problem, we treat $M_{ij}$ as a functional map~\cite{ovsjanikov2012functional} between $S_i$ and $S_j$, then perform a smooth discrete optimization algorithm~\cite{magnet2022smooth} to refine the point-to-point correspondences. We compute the point-to-point correspondences in both directions, \ie, from shape $i$ to shape $j$, denoted as $M_{ij}$, and shape $j$ to shape $i$, denoted as $M_{ji}$, then we perform a close loop check, that is, a vertex in shape $i$ mapped to $j$ using $M_{ij}$ and mapped back using $M_{ji}$ should end up in the same vertex. For $\vet{x}$ in shape $i$, we compute the mis-mapped distance as 
\eq{
\mathcal{D}_i^j = \norm{\vet{x}_i - M_{ji}(M_{ij}(\vet{x}_i))},\;. \label{eq:loop_distance}
}
We only select the correspondences that are mapped close enough, \ie, $\mathcal{D}_i^j < \delta_d$, where $\delta_d$ 
is the distance threshold to filter out mismatched vertices. Meanwhile, we compute the confidence map of correspondences using the $M_{ij}$ according to the bidirectional loop closure distance $\mathcal{D}$ in~\cref{eq:loop_distance} as 
\eq{
C_{ij} = \frac{\delta_d - D_i^j}{\max_j(D_i^j)} \;.
}

\subsection{Shape Deformation Block}
After the 3D generation and point registration block, two shapes generated from images and a set of noisy correspondences for (part of) the vertices are given. Now we need to model the deformed intermediate shapes between the pair. We solve this problem by modeling a point trajectory using a Velocity Net $\V:\mathbb{R}^3\times [0,T] \to \mathbb{R}^3$. \par
\subsubsection{Problem Formulation.}
 Given two point clouds $P_0$, $P_1 \subset \mathbb{R}^3$ sampled from two shapes $S_0$ and $S_1$, we would like to find the suitable path that not only moves $P_0$ to $P_1$, but also meets some physical condition, because the deformation between two shapes needs to be physically plausible and geometrically temporal consistent, \ie, the structural and spatial properties of the surfaces are preserved and the deformation is smooth. For point $\vet{x}\in \Omega_i \subset P_i, i\in\{0,1\}$, where $\Omega_i$ is the point domain and $\Omega = \Omega_0 \times \Omega_1$, we want to recover its trajectory $\vet{X}(t)$ with minimum Kinect energy~\cite{brenier2015optimal}. The problem can be formulated as 
\begin{align}
    \min_{\vet{X}(t)} & \int_0^T\int_{\Omega}\frac{1}{2}\norm{\frac{\dd\vet{X}(t)}{\dd t}}^2\dd\vet{x}\dd t \;, \label{eq:trajectory} \\
     \textrm{s.t.} &\int_{\Omega_0}\vet{X}(0)\dd\vet{x} = P_0, \int_{\Omega_1}\vet{X}(T)\dd\vet{x} = P_1 \;. \label{eq:overlap}
\end{align}
Instead of directly modeling the trajectory function, we model the velocity function, i.e., we estimate $\V:\mathbb{R}^3\times \mathbb{R} \to \mathbb{R}^3$, such that
\begin{equation}\label{eq:velocity}
    \V(\vet{x},t) = \frac{\dd \vet{X}(t)}{\dd t} \;,
\end{equation}
  and we could rewrite~\cref{eq:trajectory} using velocity formulation
  \begin{align} 
     \mathcal{L}_v = \int_0^T\int_{\Omega}\norm{\V(\vet{x},t)}^2 \dd\vet{x}\dd t\;. \label{eq:smooth_1} 
 \end{align}
 There are three reasons, first, to model physically plausible movement, it is easier to build physical constraints on the velocity field; second, the optimal condition in~\cref{eq:trajectory} involves velocity already. Third, the velocity field allows easy training and inferring to get the deformed shapes.
\subsubsection{Physical Plausible Velocity Field}
We need not only to solve the optimal path problem but also to ensure the path is physically realistic in the 3D world. To further control our path, we add certain constraints to our velocity field to ensure the movement generated by the velocity field is reasonable. For function smooth term in ~\cref{eq:smooth_1}, we follow~\cite{sang2025implicit} to use $\norm{\V}=\norm{(-\alpha\Delta + \vet{I})\V}$. \par
\inparagraph{Overlapping loss.} To measure how good the point cloud $P_0$ is moved by our velocity field $\V$ at time $T$, that is, our velocity field satisfies~\cref{eq:overlap}, we add one overlapping constraint as 
\eq{
\mathcal{L}_o =\dist(\int_0^T\int_{\Omega_0}\V(\vet{x},t)\dd\vet{x}\dd t - P_1) \;. \label{eq:chamfer_dist}
}
The loss describes that after $P_0$ is moved by the velocity field, how well it overlaps with $P_1$.  We use Chamfer distance as the overlapping metric in our case. \par
\inparagraph{Normal loss.} As we sample points from meshes, it is easy to access the normal of the points. We denote the normal of point $\vet{x}$ as $\vet{n}(\vet{x})$. At each time step, the point is moved by 
\eq{
\vet{x}' = \vet{x} + \V(\vet{x}, t)\Delta t \;. 
}
The local deformation of the near $\vet{x}$ can be approximated by the deformation gradient~\cite{Irgens2008}
\eq{
\vet{F}(\vet{x},t) = \vet{I} + \nabla \V(\vet{x},t)\;. \label{eq:deformation_gradient}
}
The normal update from $\vet{x}$ to $\vet{x}'$ is
\eq{
 \vet{n}' = \vet{F}^{-\top}\vet{n}\;. \label{eq:normal_update}
}

We add the normal loss to constraint the normal of point cloud $P_0$ is moved by our velocity field $\V$ should align with normal of point cloud $P_1$ at time $T$, \ie, for $\vet{x} \in \Omega_{0*}$
\eq{
\mathcal{L}_n = \norm{\int_0^T \vet{F}(\vet{x},t)^{-\top}\vet{n}(\vet{x}) \dd t - \vet{n}(\vet{x}_1)}\;, 
}

where $\Omega_0* \subset \Omega$ is the points where the correspondences are known. \par
\inparagraph{Stretching loss.} After we establish the normal update equation~\cref{eq:normal_update}, we follow the idea presented in~\cite{sang20254deform} to compute the stretching loss. The difference is, that our normal is passed through the deformation gradient operator $\vet{F}$. We define the tangent projection operator as $\vet{P} = \vet{I}-\vet{n}^\top\vet{n}$, and 
\eq{
\mathcal{L}_s = \int_{\Omega_{0}}\norm{\vet{P}^\top(\vet{F}^\top\vet{F}-\vet{I})\vet{P}}_{F}\dd \vet{x}\;,
}
where $\norm{\cdot}_F$ is the Frobenius norm~\cite{golub1996cf} of a matrix.\par
\inparagraph{Confidence Guided Matching.} Besides the total point cloud matching in~\cref{eq:chamfer_dist}, we have obtained a set of noisy correspondences and their confidence values. We use them to supervise the velocity $\V$ as well. We use the idea from the Reweighted-Least Square (RLS) Algorithm. Let $\rho(\vet{x})$ be a certain robust estimator~\cite{hampel1986robust}, and during optimization, one could plugin in weight term $w(\theta) = \frac{\dd \rho}{\dd \vet{x}}$ to perform robust optimization using loss, for $\vet{x}\in\Omega_{0*}$
\eq{
\mathcal{L}_m = w(\vet{x}_0, \vet{x}_1)\norm{\vet{x}_0 + \int_0^T\V(\vet{x},t)\dd t - \vet{x}_1}^2\;.
}
Here our confidence map is served as the robust weight function $w(\vet{x}_0, \vet{x}_1)=C_{ij}$. \par
\inparagraph{Distortion loss.} For furth control the velocity, we follow~\cite{sang20254deform} to incorporate distortion loss, for $\vet{D} = \frac{1}{2}(\nabla \mathcal{V} + (\nabla \mathcal{V})^\top)$
\eq{\label{eq:distortion}
\mathcal{L}_{d} = \int_{\dom} \norm{\frac{1}{6}\tr(\vet{D})^2-\frac{1}{2} \tr(\vet{D} \cdot \vet{D})^2 }_{F}\dd \vet{x}\;.
}

\subsection{Training}
 \inparagraph{Training.} Given a pair of keyframe images $I_0$, $I_1$, we first generate 3D meshes $S_0$, $S_1$ for each of them and the per-vertex features. Then we compute the pairwise correspondences and the confidence map. Then we sample $k=10,000$ points from meshes together with mesh vertices to form training point clouds $P_0$, $P_1$. The total loss is
 \eq{
 \label{eq:total_loss}
 \mathcal{L} = & \lambda_v \mathcal{L}_v+
 \lambda_o \mathcal{L}_o + \lambda_n \mathcal{L}_n \\ 
 & + \lambda_s \mathcal{L}_s + \lambda_m \mathcal{L}_m + \lambda_d \mathcal{L}_d \;,
 }
 where $\lambda_v$, $\lambda_o$, $\lambda_n$, $\lambda_s$, $\lambda_m$, and $\lambda_{d} \in \mathbb{R}$ are weights for each loss term.

 \par
 \inparagraph{Inferring.}  During inference time, other than directly deforming the $S_0$ vertices to generate continuous deformed mesh sequences, we can achieve also the following: \begin{enumerate*}[label=(\roman*)] 
 \item Recover dense correspondences, i.e., tracked point clouds for intermediates shape by Euler steps
 $P_i^k = P_i + \V(P_i,k)k/T $, for $k\in [0,T]$.
 \item Even during training we set $\Delta t = 1/T$ to deform the point cloud for $T$ steps and compute the losses, during inference time, we can infer on $T'\neq T$ to get lower or higher frame rates mesh sequences. 
 
 \end{enumerate*}

%% file: sec/4_experiment.tex
\section{Experiments}
\label{sec:exp}
\subsection{Setting}
\inparagraph{Network architecture.} We utilize Hunyuan3D~\cite{hunyuan3d2025} as the 3D generation backbone to obtain reconstructed meshes of the keyframes. Then, we pass the generated mesh pair to Diff3F~\cite{dutt2024diff} to get per-vertex features. Our Velocity-Net contains only an 8-layer MLP with 256 nodes, enabling fast training and near real-time inferring. Our method is implemented using purely JAX~\cite{jax2018github}.\par

\inparagraph{Datasets.} We validate our method on two types of data. For quantitative evaluation, we use \textbf{4D-DRESS}~\cite{wang20244ddress}, a real-scanned human motions dataset that has high frame rate RGB image sequences and ground truth textured 3D mesh sequences. We pick every 5th image as an input image to generate the deformed 4D sequences and compare them against the ground truth intermediate shapes.
To demonstrate the generality of our method, we also report qualitative results obtained from web images. \par
\inparagraph{Metrics.} To evaluate the 4D sequence quality and physical plausibility, we compute the Chamfer Distance (CD), and Hausdorff Distance (HD) of deformed mesh sequences using \textbf{4D-DRESS}~\cite{wang20244ddress}. We also adopt the surface area standard deviation metric \SAstd \ from 4Deform~\cite{sang20254deform} to evaluate the surface area changing among generated meshes. \par

\inparagraph{Baselines.} As we are the first method to address 4D generation for pairs of images, we propose a set of different competitive baselines. First, we consider the deformation as 2D image morphing. \textbf{DiffMorpher}~\cite{zhang2023diffmorpher} and \textbf{DreamMover}~\cite{shen2024dreammover} are state-of-the-art methods which take a source image $I_0$ and a target image $I_1$ and morph them to generate an intermediate image sequence. Both methods leverage pre-trained text-to-image diffusion models (Stable Diffusion ~\cite{rombach2022high}) and train LoRAs ~\cite{hu2022lora} to fine-tune. We treat the generated image sequence as a time-dependent 2D sequence, and we plug the intermediate frames into Hunyuan3D~\cite{hunyuan3d2025}, obtaining a 4D sequence. We also consider the ground truth intermediate 2D frames and directly reconstruct the shapes with Hunyuan3D~\cite{hunyuan3d2025}. We call such baseline \textbf{GT images}. Finally, we also apply the recent 3D-to-4D approach \textbf{4Deform}~\cite{sang20254deform}, which relies on high-quality correspondence, making it less practical. 4Deform~\cite{sang20254deform} uses the same point cloud and correspondence of TwoSquared, letting us compare the shape-deformation block. \\
\inparagraph{Training and Generation Time.} Hunyuan3D~\cite{hunyuan3d2025} requires approximately 3 minutes to generate a single 3D shape from an image. In contrast, image-morphing-based methods have a linearly increasing generation time as the desired frame rate increases. Additionally, if a different frame rate is required, the entire process must be repeated.
Our is more efficient and flexible. After the 3D generation step, the registration block takes around 3 minutes, while the shape deformation block completes training in under 1 minute. Unlike morphing-based approaches, our method can generate 4D sequences at arbitrary frame rates without additional retraining. We refer to the supplementary material~\cref{sec:supp_training} for further details on the training times.

\subsection{Validation}
In this section, we validate our method on \textbf{4D-DRESS}, where the intermediate real-world photos and scan meshes are available, and compute the quantitative metrics to demonstrate our method outperforms the other methods. 
To show the effects of each part of our loss, we perform the ablation study on the losses that are proposed by our method.\par

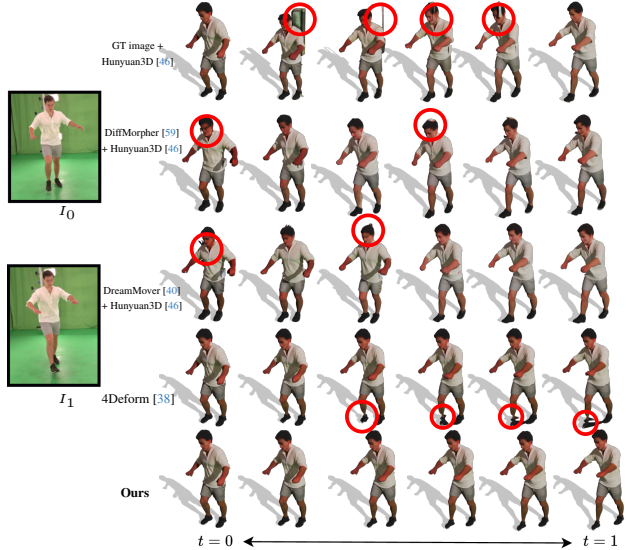
\begin{figure}
\centering
	\begin{overpic}[trim=0cm 0cm 0cm 0cm,clip, width=\linewidth]{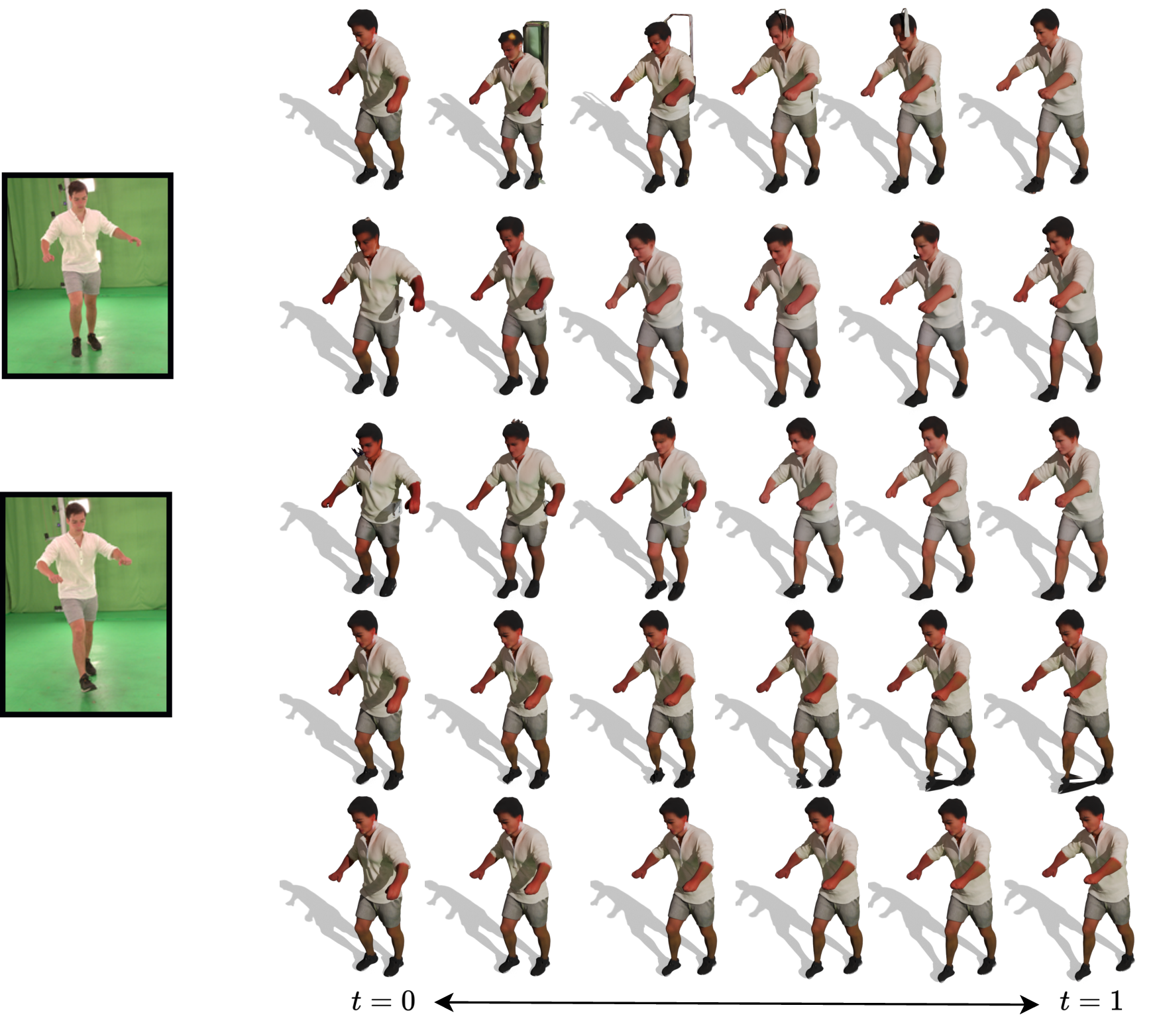}
    
		\put(15,79){\tiny{\scalebox{0.8}{\shortstack{GT image + \\
         Hunyuan3D~\cite{hunyuan3d2025}}}}}
         \put(15,65){\tiny{\scalebox{0.8}{\shortstack{DiffMorpher~\cite{zhang2023diffmorpher}\\ +
         Hunyuan3D~\cite{hunyuan3d2025}}}}}
          \put(15,40){\tiny{\scalebox{0.8}{\shortstack{DreamMover~\cite{shen2024dreammover}\\ + 
         Hunyuan3D~\cite{hunyuan3d2025}}}}}
           \put(15,25){\tiny{4Deform~\cite{sang20254deform}}}
           \put(18,10){\textbf{\tiny{Ours}}}
           \put(8,55){\tiny{$I_0$}}
           \put(8,25){\tiny{$I_1$}}

        \begin{tikzpicture}[overlay, remember picture, x=2.35pt, y=2.35pt]
            \draw[red, line width=0.5mm] (47,87) circle [radius=0.2cm];
            \draw[red, line width=0.5mm] (60,87) circle [radius=0.2cm];
            \draw[red, line width=0.5mm] (79,87) circle [radius=0.2cm];
            \draw[red, line width=0.5mm] (69,87) circle [radius=0.2cm];
            
            \draw[red, line width=0.5mm] (32,69) circle [radius=0.2cm];
            \draw[red, line width=0.5mm] (68,70) circle [radius=0.2cm];
            
            \draw[red, line width=0.5mm] (32,50) circle [radius=0.2cm];
            \draw[red, line width=0.5mm] (58,53) circle [radius=0.2cm];
            
            \draw[red, line width=0.5mm] (57,23) circle [radius=0.2cm];
            \draw[red, line width=0.5mm] (70,23) circle [radius=0.15cm];
            \draw[red, line width=0.5mm] (81,23) circle [radius=0.15cm];
            \draw[red, line width=0.5mm] (93,22) circle [radius=0.15cm];%
           \end{tikzpicture}

            \end{overpic}
\caption{\label{fig:4d_dress}\textbf{Comparison with other methods:} \methodname generates texture-consistent, physically plausible 4D sequences, and it is more robust then 4Deform~\cite{sang20254deform} to correspondences noise. In contrast, other methods show artifacts in the intermediate shapes.}
\vspace{-0.5cm}
\end{figure}

\inparagraph{Quantitative Comparison.} To quantitatively evaluate our method, we extract five keyframes from two sequences in the 4D-DRESS~\cite{wang20244ddress} dataset and generate four 4D deformation sequences using different methods. We then compute errors using the ground truth intermediate meshes. As shown in ~\cref{tab:comparison}, our method achieves lower error rates compared to other approaches and even outperforms results generated directly from ground truth intermediate images.
This suggests that fully relying on Hunyuan3D~\cite{hunyuan3d2025} would not be possible due to its inconsistent generation. In contrast, our approach ensures that as long as the starting and ending meshes are generated without significant artifacts, the intermediate shapes remain physically plausible. 
As shown in ~\cref{fig:4d_dress}, most methods can generate high-quality starting and ending meshes. However, methods that require per-step mesh generation often suffer from texture inconsistencies. We also remark that previous approaches have computational limitations, as adjusting the 4D sequence frame rate requires regenerating morphing images and re-generating meshes for each frame, making the process inefficient.
Compared to the previous 3D-to-4D generation method 4Deform~\cite{sang20254deform}, our approach is significantly more robust to noisy correspondences, leading to more consistent and high-quality 4D sequences.

\par
\input{tex/table_comparison}

\begin{figure}[ht]
	\centering
	\begin{overpic}[trim=0cm 0cm 0cm 0cm,clip, width=\linewidth]{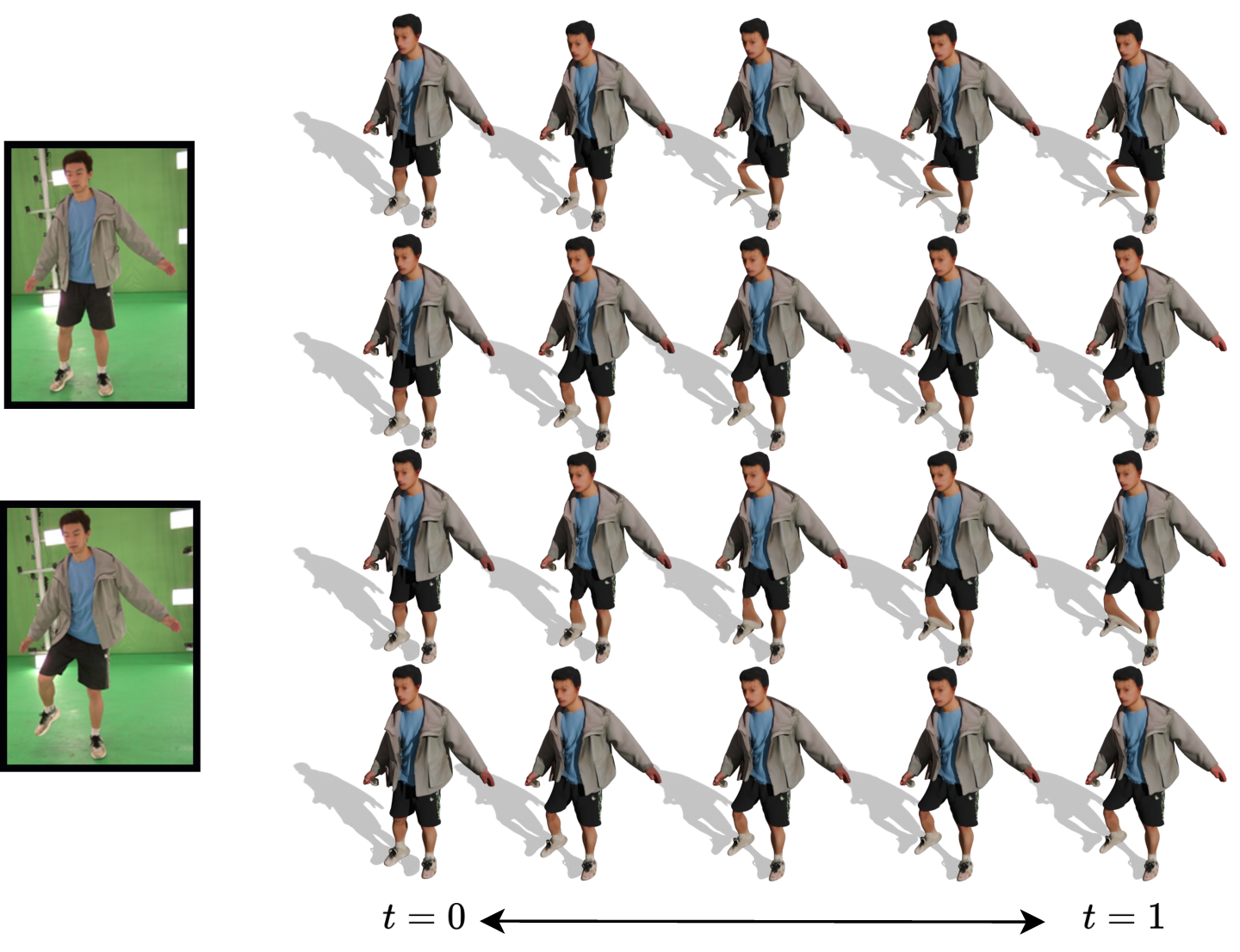}
			 \put(17,68){\tiny{w/o $\mathcal{L}_o$}}
             \put(17,48){\tiny{w/o $\mathcal{L}_n$}}
              \put(17,28){\tiny{w/o $\mathcal{L}_s$}}
            \put(17,10){\textbf{\tiny{Ours}}}
           \put(5, 41){\tiny{$I_0$}}
           \put(5,11){\tiny{$I_1$}}
			
            \put(71,23){\textcolor{red}{{\fontsize{20}{15}\selectfont {\scalebox{2}{$\circ$}}}}}
            \put(86,22){\textcolor{red}{{\fontsize{20}{15}\selectfont {\scalebox{2}{$\circ$}}}}}
            
            \put(55,57){\textcolor{red}{{\fontsize{20}{15}\selectfont {\scalebox{2}{$\circ$}}}}}
             \put(71,57){\textcolor{red}{{\fontsize{20}{15}\selectfont {\scalebox{2}{$\circ$}}}}}
            \put(87,57){\textcolor{red}{{\fontsize{20}{15}\selectfont {\scalebox{2}{$\circ$}}}}}
             
            \end{overpic}
	\caption{\textbf{Visualization of ablations: } We show the qualitative results of our ablation study. while the visual results coincide with the error number reported in~\cref{tab:ablation}, the stretching loss and overlapping loss help the shape to remain physically plausible. While the normal loss has little qualitative impact, it does lead to quantitative improvements -- see Table \ref{tab:ablation}.
	\label{fig:ablation}}
 	\vspace{-0.3cm}
\end{figure}
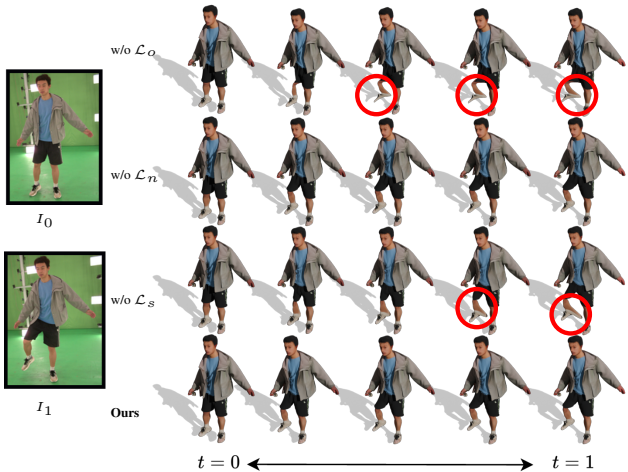

\inparagraph{Ablations.} To demonstrate that the proposed physically plausible constraint helps with generating realistic meshes, we perform a quantitative ablation for the proposed losses on \textbf{4D-DRESS}. ~\cref{tab:ablation} shows the error metrics on different settings. The overlapping loss $\mathcal{L}_o$ enforces global alignment between the deformed point cloud and the target shape, making it particularly beneficial in cases where correspondences are sparse or entirely absent in certain regions. The normal loss $\mathcal{L}_n$ ensures that the surface normals of the deformed point cloud align with those of the target. However, since normal alignment can only be enforced where correspondences exist, its effectiveness is contingent upon the quality of the estimated correspondences. As observed in ~\cref{tab:ablation}, the impact of $\mathcal{L}_n$ appears marginal in certain cases, which we attribute to suboptimal correspondence quality.
The stretching loss $\mathcal{L}_s$ generally enhances the overall quality of the deformation results. In most real-world scenarios, correspondences are neither uniformly distributed nor perfectly accurate. Compared to synthetic data, estimated correspondences in real-world cases tend to exhibit significantly lower quality, often leading to missing correspondences in specific regions. While the spatial continuity of the velocity field is enforced through ~\cref{eq:smooth_1}, this constraint alone is sometimes insufficient to prevent local distortions. The stretching loss $\mathcal{L}_s$ complements the spatial smoothness constraint by penalizing excessive local shear and stretching. As demonstrated in ~\cref{tab:ablation}, omitting $\mathcal{L}_s$ frequently results in substantial surface area deviation (\SAstd). This effect is further corroborated by the visualizations in ~\cref{fig:ablation}, where the absence of stretching loss leads to unrealistic elongation of the leg region. In summary, both quantitative and qualitative ablation studies confirm that each loss term effectively contributes to the intended objectives. Our carefully balanced loss function design ensures robust performance across various scenarios, yielding optimal results in most cases.

\input{tex/table_ablation}

\subsection{Applications}
We aim to demonstrate the generality of \methodname, which makes it suitable to work on in-the-wild images such as frames from the web. Such robustness also enables new applications, such as editing from hand sketches or pose transfer from different subjects.  \par
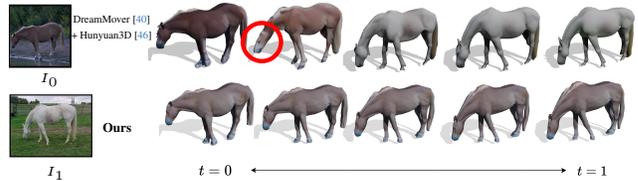
\begin{figure}[ht]
	\centering
	\begin{overpic}[trim=0cm 0cm 0cm 0cm,clip, width=\linewidth]{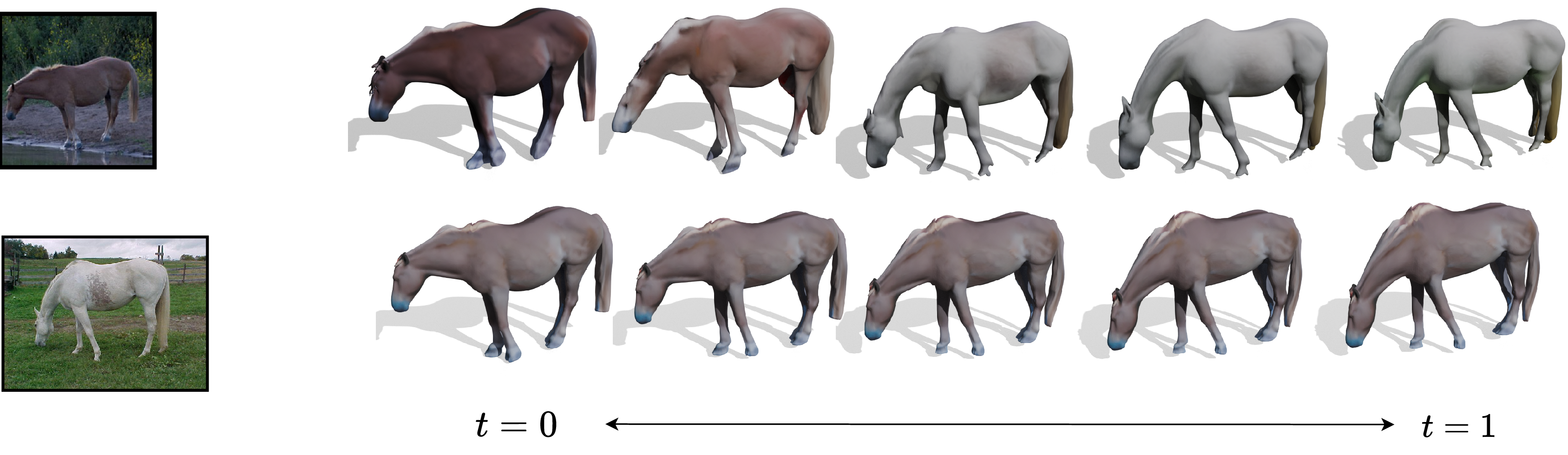}
			 \put(10,23){\tiny{\scalebox{0.8}{\shortstack{DreamMover~\cite{shen2024dreammover} \\+ 
         Hunyuan3D~\cite{hunyuan3d2025}}}}}
             \put(15,8){\textbf{\tiny{Ours}}}
           \put(5,16){\tiny{$I_0$}}
           \put(6,1){\tiny{$I_1$}}
			\put(36,18){\textcolor{red}{{\fontsize{20}{15}\selectfont {\scalebox{2}{$\circ$}}}}}
            \end{overpic}
	\caption{\textbf{4D reconstruction from web images.} Our method can take a pair of images as input and generate the temporal consistent 4D deformation between these two objects. 
	\label{fig:horse}}
\end{figure}

\inparagraph{Motion transfer from web images.} 
Our method enables the creation of dynamic 4D sequences from web images, generating smooth and realistic animations that adapt to various styles and contexts. More importantly, our approach preserves the texture of the starting image $I_0$ while adopting the gesture from the ending image $I_1$, ensuring consistency across the sequence. This means that the two input images do not need to depict the exact same object, allowing greater flexibility in choosing different sources for sequence generation.~\cref{fig:horse} shows an example of a deformation between two horse images. Our method creates temporal consistent 4D sequences, while 
DiffMorpher~\cite{zhang2023diffmorpher} fails to interpolate this image pair, and so the 3D generation task.  DreamMover~\cite{shen2024dreammover} successfully deformed the image from $I_0$ to $I_1$. However, since the following 3D generation tasks are based on the deformed 2D images, it results in texture blending. The semantic identity of some intermediate shapes can be lost as in~\cref{fig:horse}, causing physical implausibility.

\begin{figure}
\centering
	\begin{overpic}[trim=0cm 0cm 0.8cm 0cm,clip, width=\linewidth]{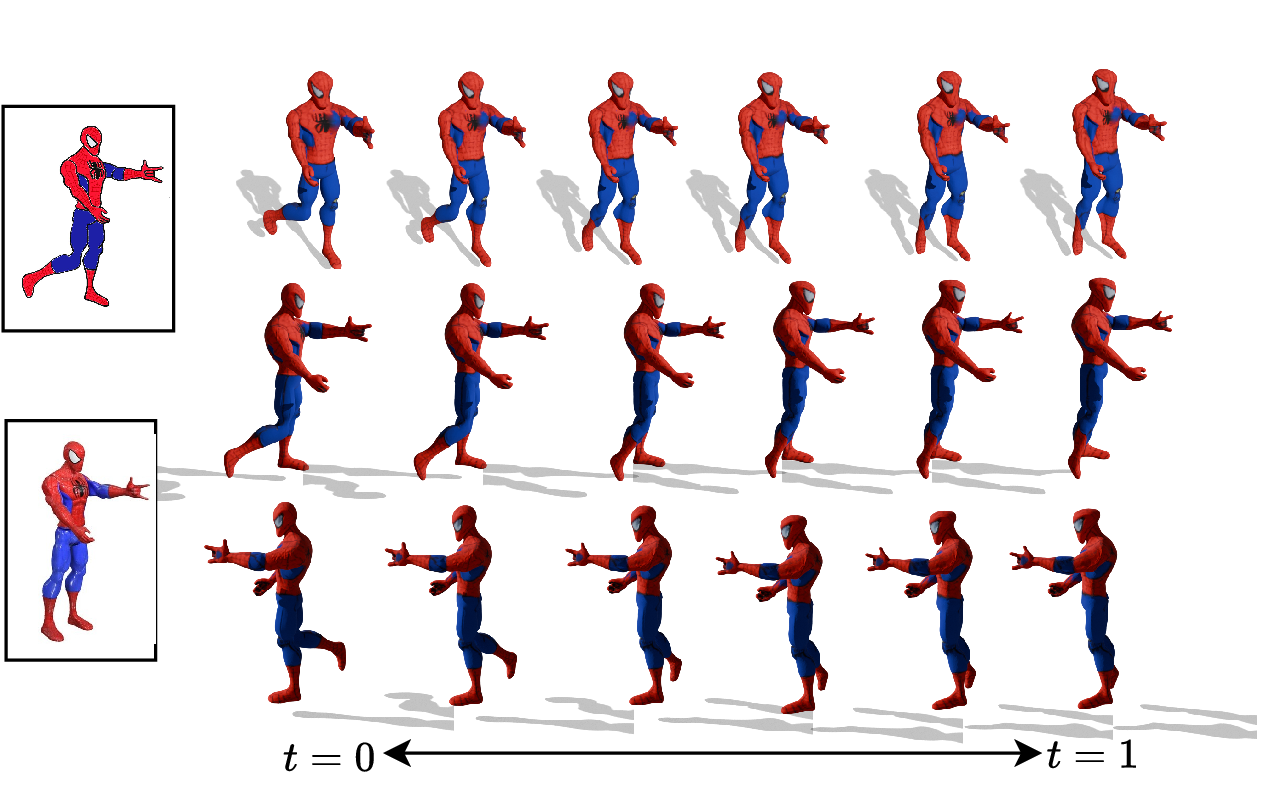}
			\put(4,38){\tiny{$I_0$}}
            \put(4, 9){\tiny{$I_1$}}

           \put(15,50){\tiny{ \shortstack{ \textbf{Ours} \\ front }}}
           \put(15,32){\tiny{ \shortstack{ \textbf{Ours} \\ side }}}
           \put(15,15){\tiny{ \shortstack{ \textbf{Ours} \\ back }}}
            \end{overpic}
\caption{\label{fig:spiderman}\textbf{Sketch images example: } Our method can take simple, low-quality images such as sketch images as input and generate high-quality meshes. }
\end{figure}
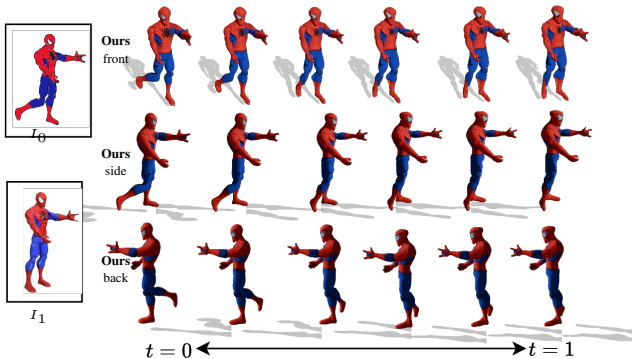

\inparagraph{4D generation from different types of images.} We demonstrate that our method accommodates a wide range of input images, extending beyond high-quality, preprocessed synthetic data. Our approach can process diverse inputs, including sketches and everyday photographs, making it adaptable to various real-world scenarios. Additionally, the objects in these images can be either synthetic models or real-world creatures. As shown in~\cref{fig:horse}, our method successfully operates on web images. Also, our method offers control over motion dynamics, providing customizable animations. We depict in~\cref{fig:spiderman} an example where the initial input image, $I_0$, is highly detailed, while the target deformation image, $I_1$, is a coarse sketch. Our method successfully generates a smooth transition from the pose in $I_0$ to the gesture in $I_1$, ensuring natural motion synthesis.

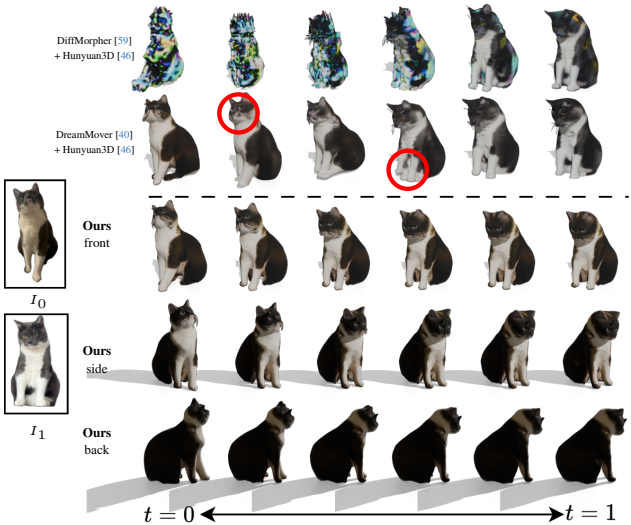
\begin{figure}
\centering
	\begin{overpic}[trim=0cm 0.1cm 0cm 0cm,clip, width=\linewidth]{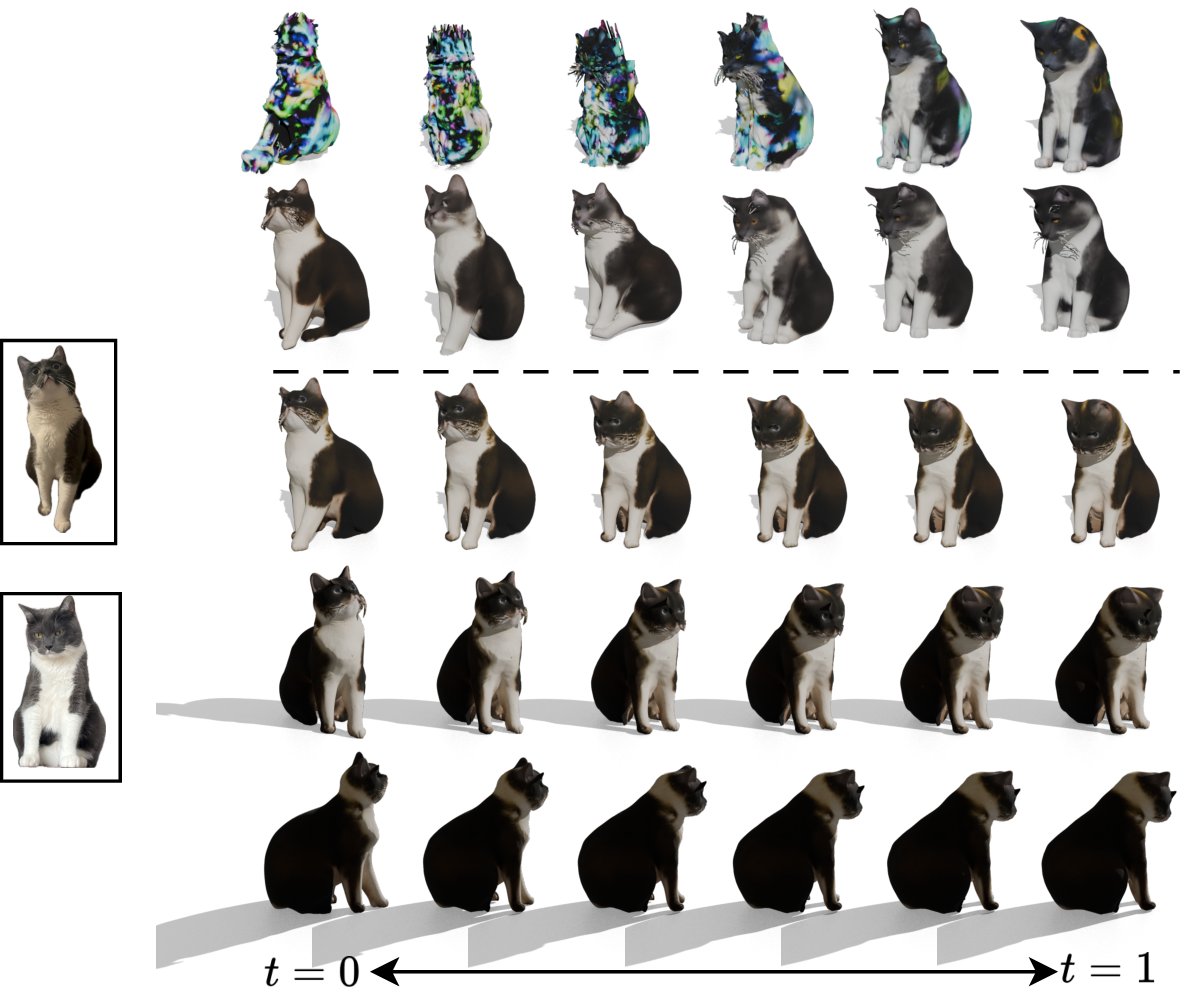}
			\put(4,36){\tiny{$I_0$}}
            \put(4, 15){\tiny{$I_1$}}
            \put(8,75){\tiny{\scalebox{0.8}{\shortstack{DiffMorpher~\cite{zhang2023diffmorpher} \\ +         Hunyuan3D~\cite{hunyuan3d2025}}}}}
          \put(8,60){\tiny{\scalebox{0.8}{\shortstack{DreamMover~\cite{shen2024dreammover}\\ +
         Hunyuan3D~\cite{hunyuan3d2025}}}}}
         \put(12,45){\tiny{ \shortstack{ \textbf{Ours} \\ front }}}
           \put(12,25){\tiny{ \shortstack{ \textbf{Ours} \\ side }}}
           \put(12,12){\tiny{ \shortstack{ \textbf{Ours} \\ back }}}

            \put(33,62){\textcolor{red}{{\fontsize{20}{15}\selectfont {\scalebox{2}{$\circ$}}}}}

            \put(60,53){\textcolor{red}{{\fontsize{20}{15}\selectfont {\scalebox{2}{$\circ$}}}}}
            \end{overpic}
\caption{\label{fig:cat}\textbf{Daily images example: } Our method can directly take images that are taken in daily life and generate continuous 4D sequences for different objects, animals, and humans. In the figure, we demonstrate the different angles of the 4D mesh sequences to show that our method generates high-quality meshes for each step.}
\end{figure}

We also show in~\cref{fig:cat} how our approach can directly process real-world images, which often contain inconsistencies such as variations in brightness. These inconsistencies lead to changes in subject appearance across multiple photos. 
Image-morphing-based 4D generation methods struggle in such cases, often failing outright or transferring such inconsistencies onto the generated meshes. Another challenge with real-world objects is their structural complexity. For example, fine details such as a cat’s whiskers are difficult to preserve using image-morphing methods (see the first red circle), which often fail to reconstruct such intricate features (see supplementary material for a comparison). In contrast, our method effectively addresses these challenges, producing 4D sequences that maintain both texture consistency and geometric integrity. To further demonstrate the quality of our results,~\cref{fig:cat} presents the generated meshes from multiple angles, highlighting the high fidelity of our reconstructions.

\subsection{Robustness Analysis and Failure Cases}
\begin{figure}
\centering
	\begin{overpic}[trim=0cm 0cm 0cm 0cm,clip, width=\linewidth]{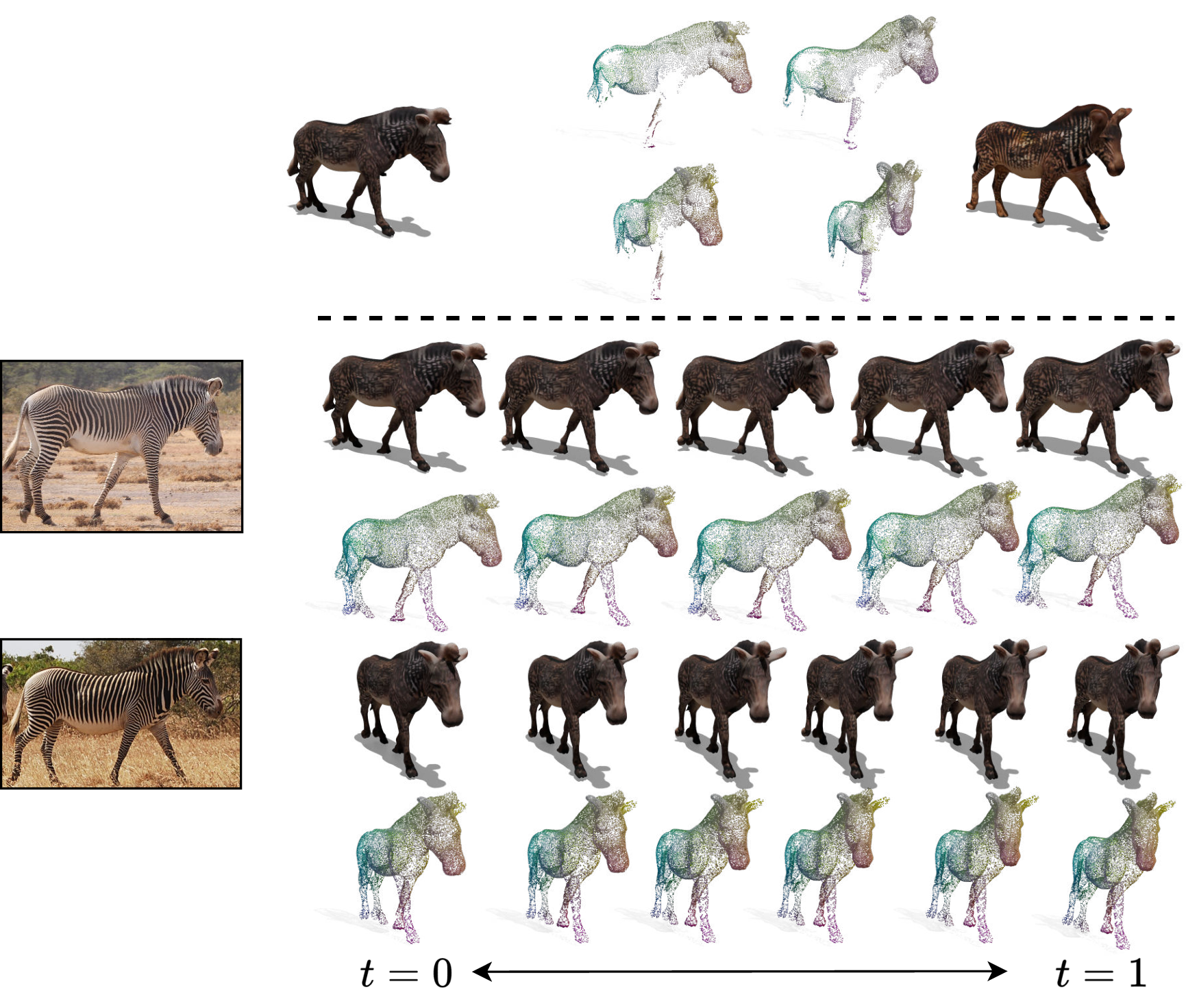}
			\put(8,37){\tiny{$I_0$}}
            \put(8, 15){\tiny{$I_1$}}

            \put(27,65){\tiny{$S_0$}}
            \put(88,65){\tiny{$S_1$}}

            \put(38,64){\tiny{Correspondences}}
            \put(36,74){\tiny{Correspondences}}

            \put(22, 51){\tiny{\shortstack{\textbf{Ours} \\ side}}}
            \put(22, 38){\tiny{\shortstack{tracked \\ points}}}

             \put(22, 24){\tiny{\shortstack{\textbf{Ours} \\front}}}
            \put(22, 12){\tiny{\shortstack{tracked \\ points}}}

             \end{overpic}
\caption{\label{fig:zebra}\textbf{Robustness analysis: }In cases where vertex registration fails, our method may not to fully deform the shape to match the target $S_1$, However, it effectively preserves the physical characteristics of regions where correspondences are missing (e.g., the legs) while successfully adapting the registered parts (e.g., the head) to align with the target shape. Despite these limitations, the intermediate shapes remain physically plausible, and texture consistency is maintained throughout the sequence. 
}
\end{figure}
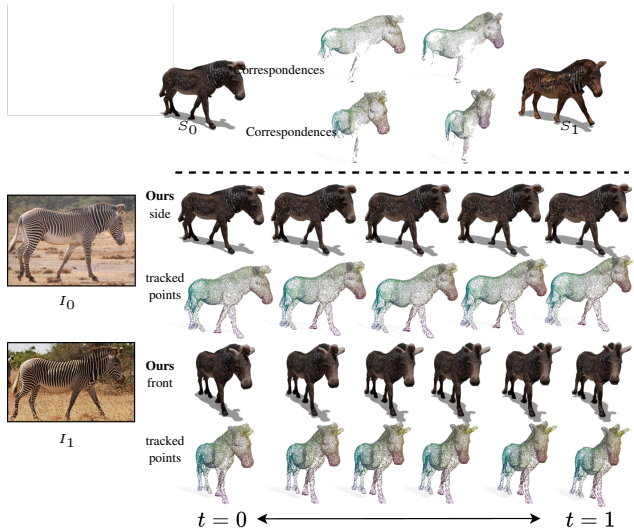

Our method takes images from different sources as input and obtains correspondences between two shapes. However, these correspondences may sometimes be inaccurate due to low-quality images or the complex geometric features of the shapes—particularly in symmetric structures. Our method demonstrates resistance to such inaccuracies. By incorporating a closed-loop check (see ~\cref{sec:method} or ~\cref{fig:pipeline}), we effectively filter out significantly misaligned correspondences. Even in cases of partial assignments, our method still produces reasonable results. ~\cref{fig:zebra} illustrates a case where correspondences are largely missing in the leg region. Despite having only one successfully registered leg, our method still generates physically plausible sequences. Rather than forcing a complete match to the target shape, our approach prioritizes physical plausibility, resulting in more stable deformations with minimal distortion. Furthermore, since our velocity field deforms the whole sampled point cloud, the points are tracked, establishing a dense correspondence throughout the sequence.

%% file: tex/table_comparison.tex
\begin{table}[t]
\footnotesize
\addtolength{\tabcolsep}{-3pt}
\begin{tabular}{llccc}
\specialrule{.1em}{.05em}{.05em} 
        \toprule
 Seq. & Method &  CD \tiny{$(\times 10^2)\downarrow$} & HD \tiny{$(\times 10)\downarrow$} & \SAstd \tiny{$(\times 10)\downarrow$}  \\ 
 \cmidrule{3-5}
\multirow{5}{*}{Take22} & GT image & 1.503 & 1.998 & \textbf{0.201} \\
&DiffMorpher~\cite{zhang2023diffmorpher} & 1.678 & 2.029 & 1.161 \\
&DreamMover~\cite{shen2024dreammover} & 1.683 & 2.033 & 1.116 \\
&4Deform~\cite{sang20254deform} & \textbf{1.434} & 1.997 & 0.217   \\
\cmidrule{2-5}
&\textbf{Ours} & 1.451 & \textbf{1.996} & \textbf{0.201} \\
\specialrule{.1em}{.05em}{.05em} 
\multirow{5}{*}{Take7} & GT image & 0.099 & 0.145 & 1.079\\
&DiffMorpher~\cite{zhang2023diffmorpher} & 0.110 & 0.710 & 1.228  \\
&DreamMover~\cite{shen2024dreammover} & 0.126 & 0.186 & 1.359 \\
&4Deform~\cite{sang20254deform} & 0.079 & 0.137& 0.173   \\
\cmidrule{2-5}
&\textbf{Ours} & \textbf{0.074} & \textbf{0.117} & \textbf{0.139} \\
\bottomrule
\end{tabular}
\caption{\textbf{Quantitative comparison:} We show that our method achieves the best quantitative results compare to previous methods.}
\label{tab:comparison}
\end{table}

%% file: tex/table_ablation.tex
\begin{table}[t]
\footnotesize
\centering
\addtolength{\tabcolsep}{-3pt}
\begin{tabular}{llccc}
\specialrule{.1em}{.05em}{.05em} 
        \toprule
 Seq. & Method &  CD \tiny{$(\times 10^2)\downarrow$} & HD \tiny{$(\times 10)\downarrow$} & \SAstd \tiny{$(\times 10)\downarrow$}  \\ 
 \cmidrule{3-5}
\multirow{4}{*}{Take22} & w/o $\mathcal{L}_o$ & 1.462 & 1.998 & \textbf{0.201}  \\
& w/o $\mathcal{L}_n$ & 1.539 & \textbf{1.965} & 0.202 \\
& w/o $\mathcal{L}_s$ & 1.458 & 2.004 & 0.223  \\
\cmidrule{2-5}
&\textbf{Ours} & \textbf{1.451} & 1.996 & \textbf{0.201} \\
\specialrule{.1em}{.05em}{.05em} 
\multirow{4}{*}{Take7} &w/o $\mathcal{L}_o$ & 0.075 & 0.119& 0.140\\
&w/o $\mathcal{L}_n$ & \textbf{0.074} & 0.119 & \textbf{0.137}  \\
&w/o $\mathcal{L}_s$ & 0.081 & 0.126 &0.486 \\
\cmidrule{2-5}
&\textbf{Ours} & \textbf{0.074} & \textbf{0.117} & 0.139 \\
\bottomrule
\end{tabular}
\caption{\textbf{Quantitative ablation:} We show how each part of our losses affects the final results. }
\label{tab:ablation}
\end{table}

%% file: sec/5_conclusion.tex
\section{Conclusion \& Future work }
\label{sec:conclusion}
We presented \methodname as the first approach that generates a complete 4D sequence of an arbitrary object from just a pair of images. \methodname combines the latest advances in 3D reconstruction with physically plausible modeling of the deformation. We also revisit the best 4D deformation module available at the present date, enabling it to work with sparse and noisy correspondence, while making it significantly faster. 
While \methodname shows compelling results on an entirely new challenge, there remain significant open problems to resolve. While we address a broad class of objects (non-rigid deformation of articulated shapes), it is unclear how our deformations model would scale to more intricate scenarios like hair, where thousands of separate deformations happen simultaneously. Such challenges would require highly precise tracking, which is not available at present date. Nevertheless, \methodname opens up the interesting new challenge of 4D reconstruction from minimal input.  We believe it can be instrumental for Generative AI pipelines, providing supervision and guidance in dynamic generations, such as videos.

\section*{Acknowledgements}
We would like to thank Tarun Yenamandra and Viktoria Ehm for helpful insights and proofreading.

%% file: sec/X_suppl.tex
\clearpage
\setcounter{page}{1}
\setcounter{section}{-1}

\renewcommand \thesection{S\arabic{section}}
\renewcommand{\thefigure}{S.\arabic{figure}}
\renewcommand{\thetable}{S.\arabic{table}}
\renewcommand{\theequation}{s.\arabic{equation}}

\twocolumn[{%
\renewcommand\twocolumn[1][]{#1}%
\maketitlesupplementary

\begin{center}
    \centering
    \begin{overpic}[trim=0cm 0.2cm 0cm 0.4cm,clip, width=\linewidth]{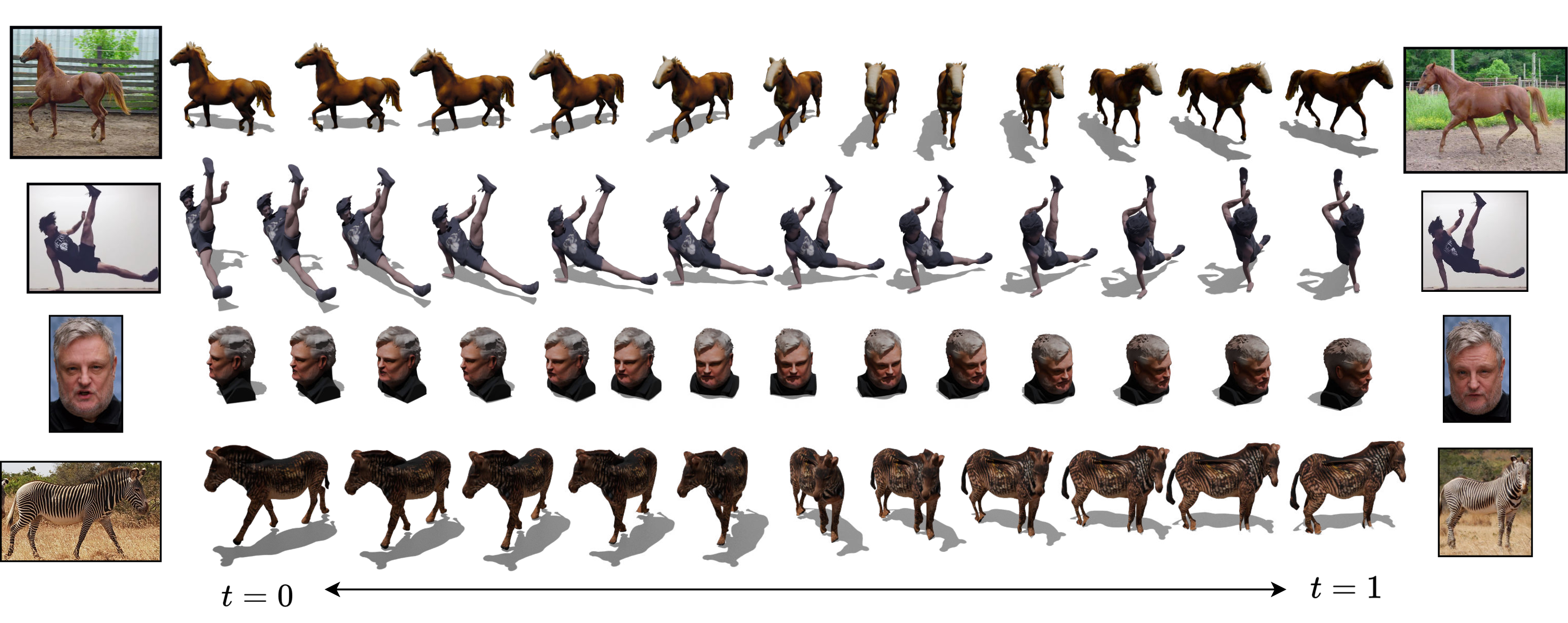}
			\put(6, 0){\small{$I_0$}}
            \put(94, 0){\small{$I_1$}}
            \end{overpic}
    \captionof{figure}{\methodname takes a pair of 2D images representing the initial and final states of an object as input and generates texture-consistent, geometry-consistent 4D continuous sequences. It is designed to be robust to varying input quality, operating without the need for predefined templates or object-class priors. This adaptability enables greater flexibility in processing diverse images while maintaining structural integrity and visual coherence throughout the generated sequences. As demonstrated, our approach effectively handles humans, animals, and inanimate objects.}\label{fig:supp_teaser}
\end{center}%

}]

\section{Training and Inference Details}\label{sec:supp_training}
\inparagraph{TwoSquared (ours).} Given a pair of images, we use Hunyuan3D~\cite{hunyuan3d2025} for obtaining their 3D reconstruction. It first encodes each image into a latent space and then decodes each code into a Signed Distance Field (SDF) to generate a textureless mesh using marching cubes. Finally, a high-resolution texture is added to the generated mesh using the texture painting module conditioned on the input image. This whole process of obtaining textured meshes takes around 3 minutes per input image. Then we use Diff3F~\cite{dutt2024diff} to compute the feature for each vertex. This procedure takes 5 minutes per frame. Then we train our Velocity Net for 4,000 epochs, which takes \emph{less than 2 minutes}.  Therefore, each pair input takes roughly 17 minutes to train. Our velocity field is continuous in time, and inference can happen at arbitrary numbers of steps without retraining. The inferring time to generate a mesh takes \emph{less than 1 second}. To infer the higher frame rates, we suggest using $T' < 2T$ to ensure satisfactory results. \par
\inparagraph{DiffMorpher~\cite{zhang2023diffmorpher}: } The morphing method takes two input images and \emph{text descriptions} to obtain a morphed sequence between source and target, taking 2 minutes to obtain the intermediate images. Then, the generated images of each timestep are fed through the 3D reconstruction network. We remark that 3D reconstruction takes 3 minutes, making this technique particularly slow at inference. In total, \textit{20 minutes} are needed to obtain deformed 3D shapes in 6 timesteps.

\inparagraph{DreamMover~\cite{shen2024dreammover}:} The image morphing method takes 5 minutes to interpolate a pair of the image. Same as DiffMorpher~\cite{zhang2023diffmorpher}, this method leverages a pre-trained text-to-image diffusion model and also requires a text prompt. The resulting sequence of images is then fed to Hunyuan3D~\cite{hunyuan3d2025}, which takes $3$ minutes for every frame. In our experiments, it requires \emph{23 minutes} in total for a sequence of 5 timesteps. 
\inparagraph{4Deform~\cite{sang20254deform}:} We give the method the point clouds from Hunyuan3D for the initial and final frame and run the same $4000$ epoch for their Velocity Net (takes less than 2 minutes) and total $10000$ epochs plus Implicit Net, which takes around $12$ minutes. It can infer arbitrary steps after training as ours, and each step takes less than 1 second.

\section{Morphed Images}
\begin{figure}[ht]
	\centering
    \hfill
	\begin{overpic}[trim=0cm 0cm 0cm 0cm, width=0.95\linewidth]{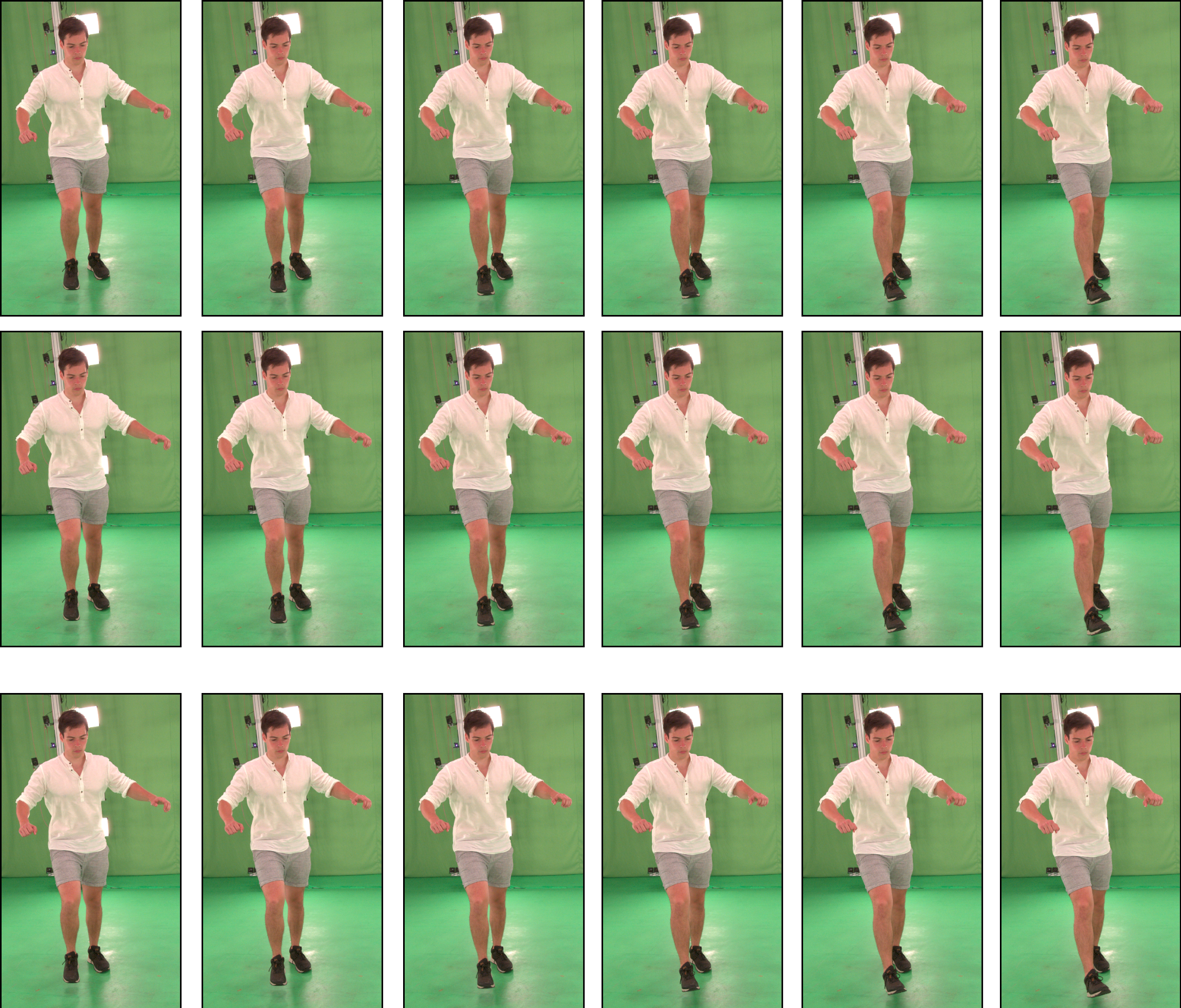}
			 \put(-5,61){\scriptsize{\rotatebox{90}{GT image}}}
         \put(-5,31){\scriptsize{\rotatebox{90}{DiffMorpher~\cite{zhang2023diffmorpher}}}}
          \put(-5,1){\scriptsize{\rotatebox{90}{DreamMover~\cite{shen2024dreammover}}}}
			
            \end{overpic}
	\caption{\textbf{Morphing results for 4D-DRESS.} We show the ground truth images and the morphed images using DiffMorpher~\cite{zhang2023diffmorpher}, DreamMover~\cite{shen2024dreammover} in~\cref{fig:4d_dress} in the main paper. These are images that are passed to the 3D generation backbone to generate intermediate shapes in~\cref{fig:4d_dress}. 
	\label{fig:sub_take7}}
 	\vspace{-0.3cm}
\end{figure}

We provide additional results to complement the experiments presented in the main paper. As shown in~\cref{fig:sub_take7}, DiffMorpher~\cite{zhang2023diffmorpher} and DreamMover~\cite{shen2024dreammover} successfully generate high-quality intermediate keyframes that are almost identical to the ground-truth images. We then use these three sequences as inputs to the 3D generation backbone to produce intermediate shapes, which are visualized in~\cref{fig:4d_dress}. While~\cref{fig:sub_take7} demonstrates that the image morphing methods can create visually plausible keyframes, it also highlights a key limitation: generating 4D sequences by independently conditioning on each keyframe often introduces artifacts and fails to maintain geometric and texture consistency.

\begin{figure}[ht]
	\centering
    \hfill
	\begin{overpic}[trim=0cm 0cm 0cm 0cm, width=0.95\linewidth]{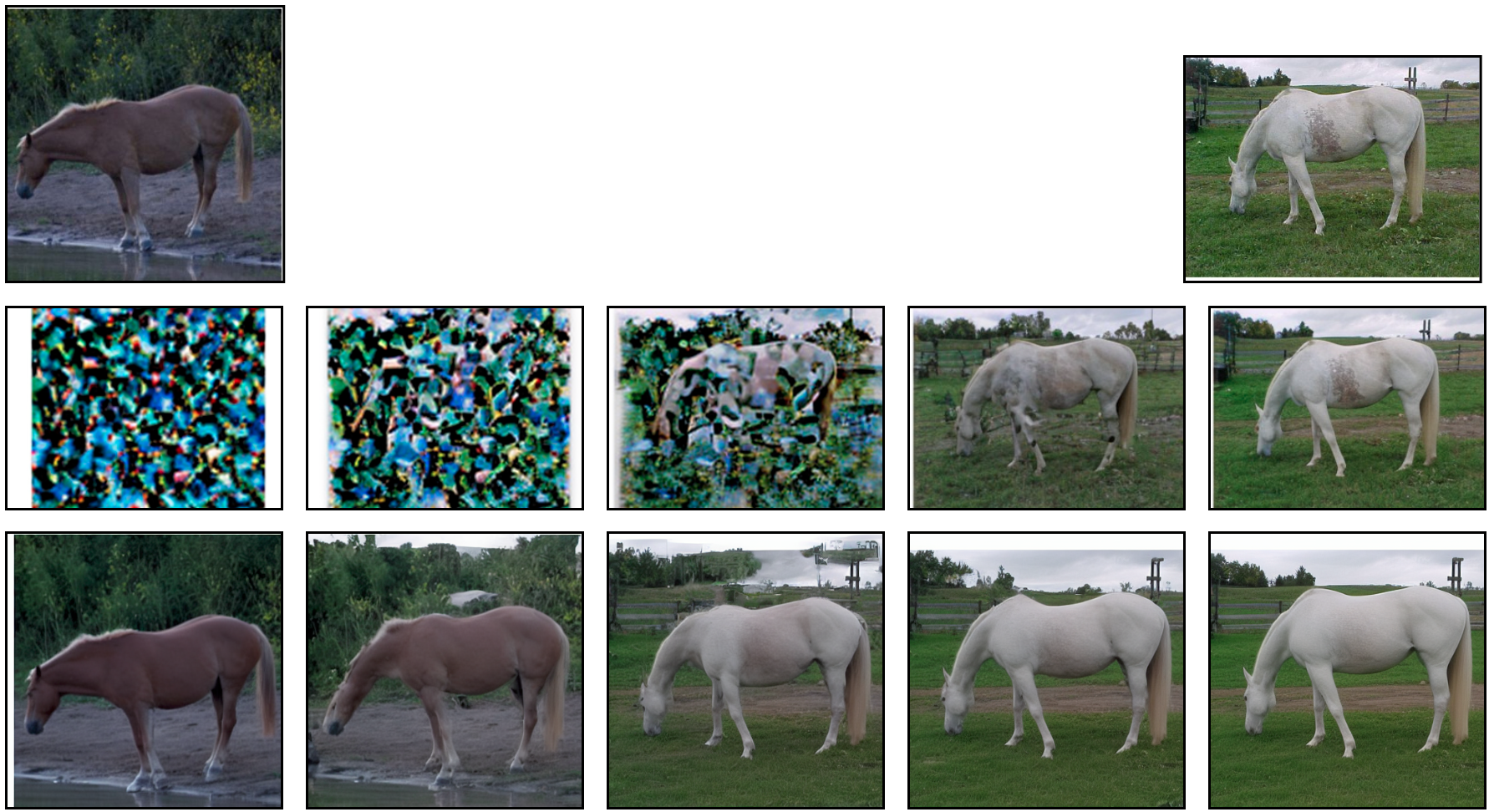}
			 \put(-5,37){\tiny{\rotatebox{90}{input image}}}
         \put(-5,18){\tiny{\rotatebox{90}{DiffMorpher~\cite{zhang2023diffmorpher}}}}
          \put(-5,0){\tiny{\rotatebox{90}{DreamMover~\cite{shen2024dreammover}}}}
            \end{overpic}
	\caption{\textbf{Morphing results for horse.} We show the ground truth images and the morphed images using DiffMorpher~\cite{zhang2023diffmorpher}, DreamMover~\cite{shen2024dreammover} in~\cref{fig:horse} in the main paper. These are images that are passed to the 3D generation backbone to generate intermediate shapes. As DiffMorpher heavily failed so the 3D generation steps also failed. 
	\label{fig:sub_horse}}
 	\vspace{-0.3cm}
\end{figure}

\begin{figure}[ht]
	\centering
	\begin{overpic}[trim=0cm 0cm 0cm 0cm, width=0.85\linewidth]{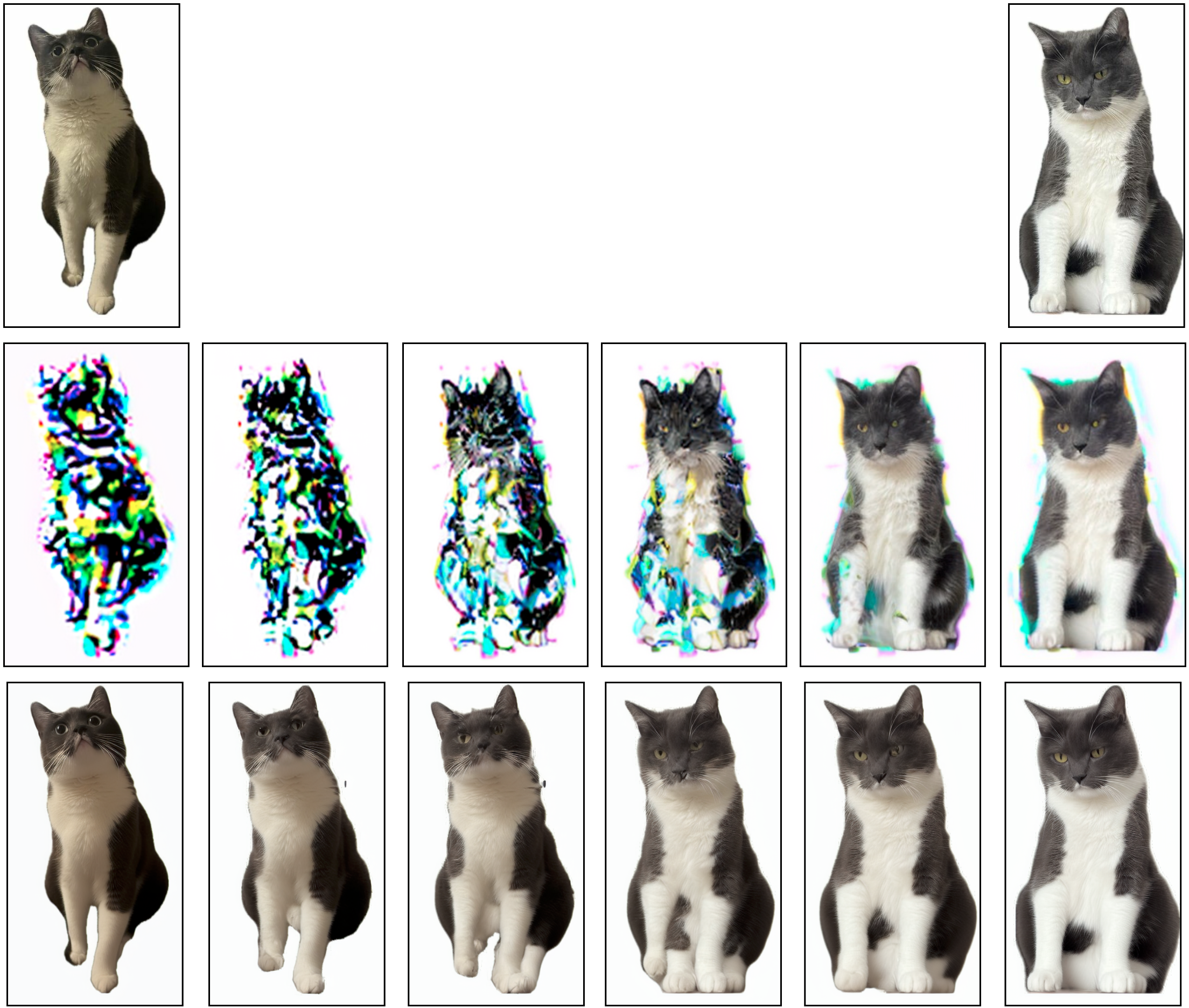}
			 \put(-5,60){\footnotesize{\rotatebox{90}{input image}}}
         \put(-5,30){\footnotesize{\rotatebox{90}{DiffMorpher~\cite{zhang2023diffmorpher}}}}
          \put(-5,0){\footnotesize{\rotatebox{90}{DreamMover~\cite{shen2024dreammover}}}}
      
            \end{overpic}
	\caption{\textbf{Morphing results for cat.} We show the ground truth images and the morphed images using DiffMorpher~\cite{zhang2023diffmorpher}, DreamMover~\cite{shen2024dreammover} in~\cref{fig:cat} in the main paper. These are images that are passed to the 3D generation backbone to generate intermediate shapes. 
	\label{fig:sub_cat}}
 	\vspace{-0.3cm}
\end{figure}

We also present the morphed images corresponding to~\cref{fig:horse} in the main paper. As shown in~\cref{fig:sub_horse}, DiffMorpher~\cite{zhang2023diffmorpher} failed to interpolate between the two different horse images, leading to subsequent failures in the 3D generation process.
DreamMover~\cite{shen2024dreammover} produces reasonable results by interpolating between images. However, since its task is designed to match the start and end images, it naturally interpolates not only the shape but also the changing texture. While this works well for objects that remain identical, it introduces texture inconsistency when handling two different objects of the same species, ultimately affecting the 3D generation output. This suggests that if 4D generation relies purely on image interpolation, the start and end images must represent the exact same object to maintain consistency.
Moreover, a closer examination of DreamMover’s results in~\cref{fig:sub_horse} reveals that the horse's movement is not temporally smooth across the morphed images, and the interpolation fails to maintain physical realism. The same situation happens to the cat example as shown in~\cref{fig:sub_cat}. See in~\cref{fig:sub_cat} the cat leg's motion is not continuous and smooth.

\section{Additional Visualizations}

\begin{figure}[ht]
	\centering
	\begin{overpic}[trim=0cm 0.4cm 1.2cm 1.2cm, width=\linewidth]{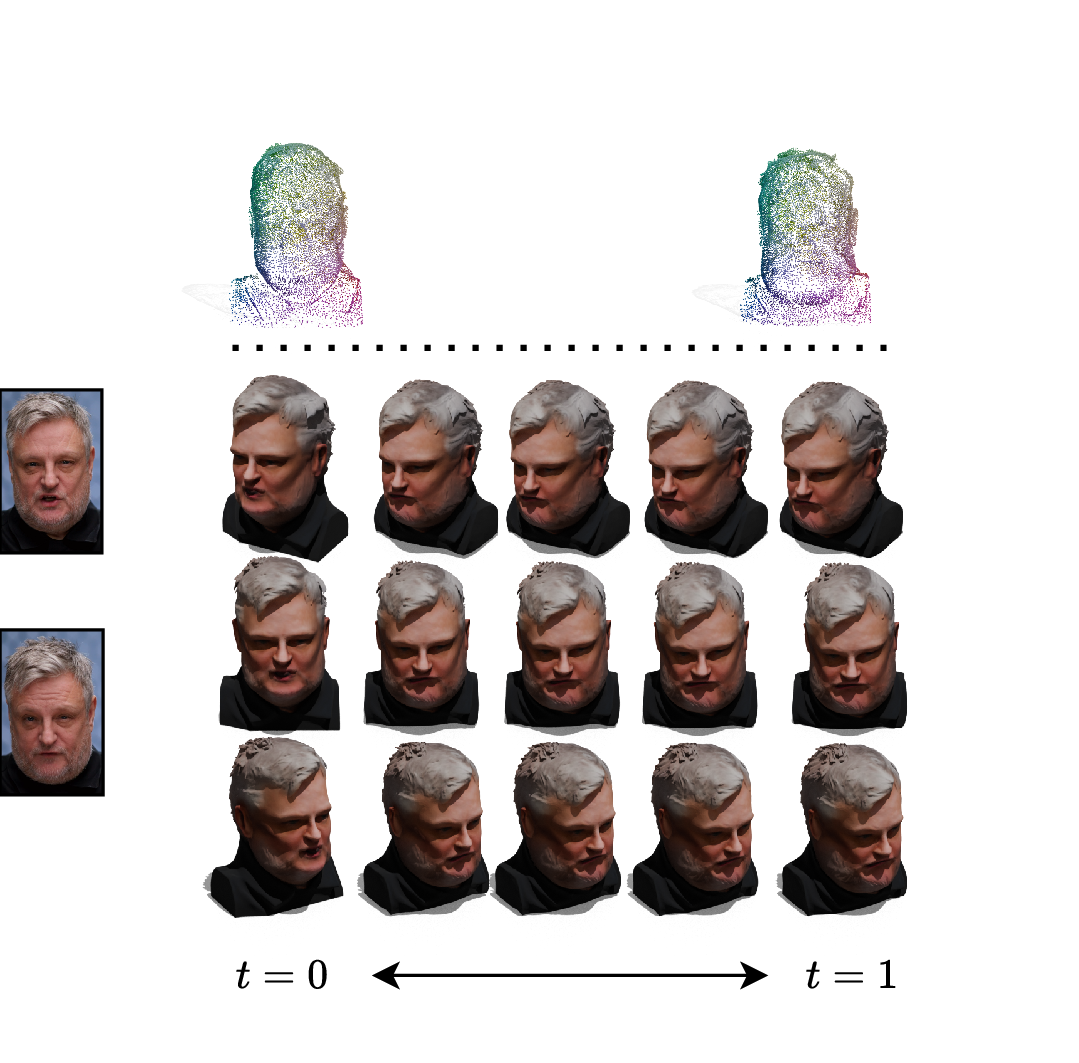}

            \put(5,45){\footnotesize{$I_0$}}
            \put(5, 19){\footnotesize{$I_1$}}

			\put(3,80){\footnotesize{Correspondences}}

            \put(12, 51){\footnotesize{\shortstack{\textbf{Ours} \\ left}}}

             \put(12, 33){\footnotesize{\shortstack{\textbf{Ours} \\front}}}
              \put(12, 18){\footnotesize{\shortstack{\textbf{Ours} \\right}}}
            
            \end{overpic}
	\caption{\textbf{Visualization from different angle.} We show our 4D sequences by showing the deformed mesh sequence at different angles. The correspondences on the human face are noisy, however, our method still gets reasonable deformations.
	\label{fig:oldman}}
 	\vspace{-0.3cm}
\end{figure}

In this section, we add more visualization results to demonstrate our method together with our estimated correspondences. Showcase that our method is robust to noisy and incomplete correspondence situations. We also demonstrate that our method can handle different types of objects, including Some types that are used to be solved using a template, such as human face~\cref{fig:oldman}, and animals~\cref{fig:supp_teaser}. Compared to the template-based method, our method is more faithful to the real-world details of the objects.
\begin{figure}[ht]
	\centering
	\begin{overpic}[trim=0cm 0cm 0.1cm 0.2cm, width=\linewidth]{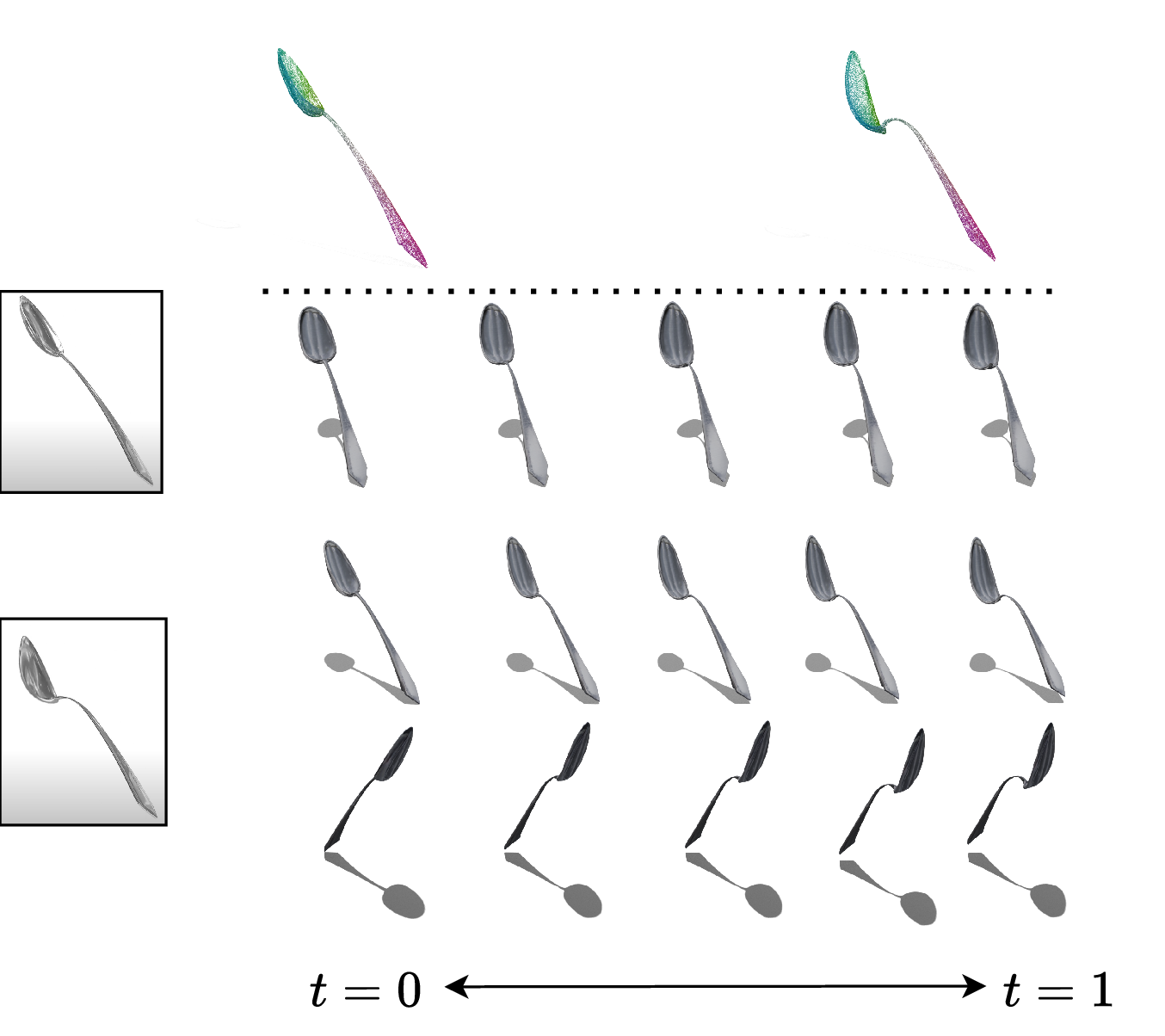}

            \put(5,40){\footnotesize{$I_0$}}
            \put(5, 13){\footnotesize{$I_1$}}

			\put(8,72){\footnotesize{Correspondences}}

            \put(20, 51){\footnotesize{\shortstack{\textbf{Ours} \\ front}}}

             \put(20, 33){\footnotesize{\shortstack{\textbf{Ours} \\side}}}
              \put(20, 18){\footnotesize{\shortstack{\textbf{Ours} \\back}}}
            
            \end{overpic}
	\caption{\textbf{Visualization from different angle.} We show our 4D sequences by showing the deformed mesh sequence at different angles.
	\label{fig:spoon}}
 	\vspace{-0.3cm}
\end{figure}

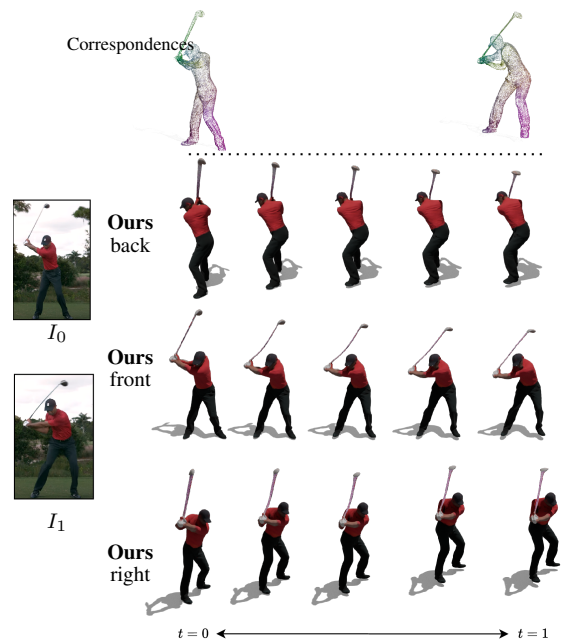
\begin{figure}[ht]
	\centering
	\begin{overpic}[trim=0cm 0.4cm 1.4cm 1.2cm, width=\linewidth]{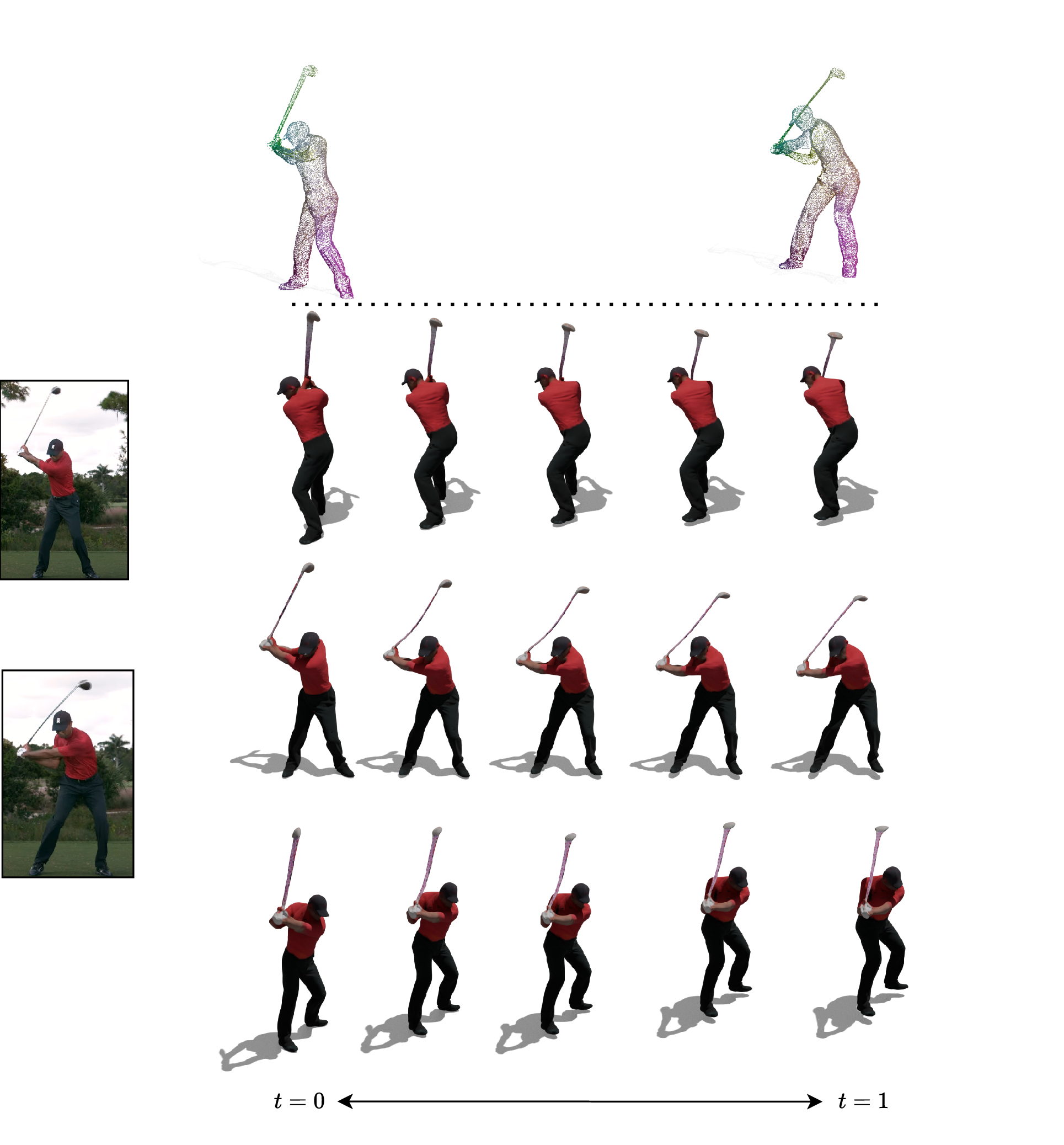}

            \put(5,47){\footnotesize{$I_0$}}
            \put(5, 19){\footnotesize{$I_1$}}

			\put(8,90){\scriptsize{Correspondences}}

            \put(14, 60){\footnotesize{\shortstack{\textbf{Ours} \\ back}}}

             \put(14, 40){\footnotesize{\shortstack{\textbf{Ours} \\front}}}
              \put(14, 11){\footnotesize{\shortstack{\textbf{Ours} \\right}}}
            
            \end{overpic}
	\caption{\textbf{Visualization from different angle.} We show our 4D sequences by showing the deformed mesh sequence at different angles. The correspondence after the refinement is smooth and accurate. 
	\label{fig:tiger}}
 	\vspace{-0.3cm}
\end{figure}